\documentclass{article}
\usepackage{tikz}
\usetikzlibrary{positioning, arrows.meta}

\usepackage{microtype}
\usepackage{graphicx}
\usepackage{subfigure}
\usepackage{subcaption}
\usepackage{booktabs} % for professional tables
\usepackage[table]{xcolor}
\usepackage{overpic}

\usepackage{hyperref}
\usepackage{enumitem}

\makeatletter
\providecommand{\Notice@String}{}
\makeatother

\usepackage[accepted]{icml2025}

\usepackage{amsmath}
\usepackage{amssymb}
\usepackage{mathtools}
\usepackage{amsthm}

\usepackage[capitalize,noabbrev]{cleveref}

\theoremstyle{plain}

\theoremstyle{definition}

\theoremstyle{remark}

\usepackage[textsize=tiny]{todonotes}

\icmltitlerunning{Layerwise Progressive Freezing Enables STE-Free Training of Deep Binary Neural Networks}

\begin{document}

\twocolumn[
\icmltitle{Layerwise Progressive Freezing Enables STE-Free Training of Deep Binary Neural Networks}



\icmlsetsymbol{equal}{*}

\begin{icmlauthorlist}
\icmlauthor{Evan Gibson Smith}{WPI}
\icmlauthor{Bashima Islam}{WPI}
\end{icmlauthorlist}

\icmlaffiliation{WPI}{Worcester Polytechnic Institute}

\icmlcorrespondingauthor{Evan Gibson Smith}{egsmith@wpi.edu}

\icmlkeywords{Binary Neural Networks, Network Quantization, Progressive Freezing, Layerwise Training, Model Compression, Machine Learning, ICML}

\vskip 0.3in
]



\printAffiliationsAndNotice{}  % leave blank if no need to mention equal contribution
\newcommand\bnote[1]{\textcolor{blue}{#1}}
\newcommand\rnote[1]{\textcolor{red}{#1}}

\begin{abstract}
    We investigate progressive freezing as an alternative to straight-through estimators (STE) for training binary networks from scratch. Under controlled training conditions, we find that while global progressive freezing works for binary-weight networks, it fails for full binary neural networks due to activation-induced gradient blockades. We introduce StoMPP (Stochastic Masked Partial Progressive Binarization), which uses layerwise stochastic masking to progressively replace differentiable clipped weights/activations with hard binary step functions, while only backpropagating through the unfrozen (clipped) subset (i.e., no straight-through estimator). Under a matched minimal training recipe, StoMPP improves accuracy over a BinaryConnect-style STE baseline, with gains that increase with depth (e.g., for ResNet-50 BNN: +18.0 on CIFAR-10, +13.5 on CIFAR-100, and +3.8 on ImageNet; for ResNet-18: +3.1, +4.7, and +1.3). For binary-weight networks, StoMPP achieves 91.2\% accuracy on CIFAR-10 and 69.5\% on CIFAR-100 with ResNet-50. We analyze training dynamics under progressive freezing, revealing non-monotonic convergence and improved depth scaling under binarization constraints.
\end{abstract}
\section{Introduction}
Quantized neural networks promise reduced memory footprint and faster, more energy-efficient inference, but training them remains challenging: optimization proceeds in a continuous parameter space, while deployment constrains parameters (and sometimes activations) to a discrete set. We focus on the extreme end of this spectrum, \emph{binary} networks, where weights and/or activations take values in $\{-1,+1\}$. Binary-weight networks (BWNs) and full binary neural networks (BNNs) can substantially reduce bandwidth and arithmetic cost, but often suffer from accuracy degradation and optimization instability.

Most successful approaches to training BWNs and BNNs rely on straight-through estimator (STE) style updates, popularized by BinaryConnect and subsequent BNN training methods \cite{BinaryConnect_Courbariaux_2016}. These methods use a hard binarization operator (e.g., $\mathrm{sign}$) in the forward pass and replace its derivative in the backward pass with a surrogate. While empirically effective, this introduces a mismatch between the executed forward operator and the gradient used for learning, and the mismatch can become more pronounced as networks deepen. Prior work improves BNN performance through refined approximations, optimizers, and architectural modifications \cite{ReCU_Xu_2021, BiPer_Vargas_2024a, OvSW_Xiang, XNOR_Net_Rastegari_2016, BiReal_Liu_2018, ReActNet_Liu_2020}, but most gradient-based training pipelines still depend on STE-like gradient replacements in some form.

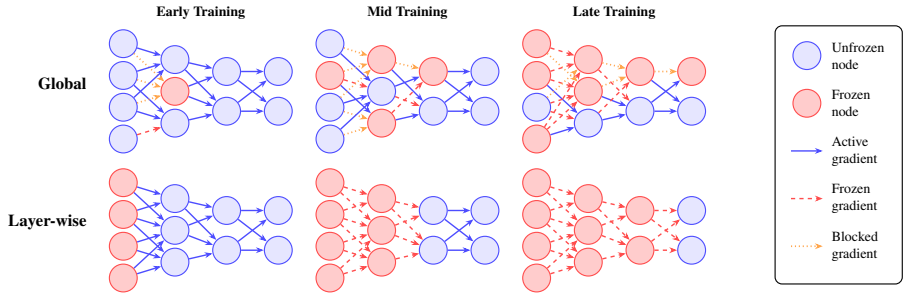
\begin{figure*}[t]
\centering

% Define styles with color
\tikzstyle{unfrozen_node} = [circle, draw=blue!70, fill=blue!10, thick, minimum size=7mm]
\tikzstyle{frozen_node} = [circle, draw=red!70, fill=red!20, thick, minimum size=7mm]
\tikzstyle{active_edge} = [-{Stealth[length=2mm]}, thick, blue!70, line width=1pt]
\tikzstyle{frozen_edge} = [-{Stealth[length=2mm]}, thick, red!70, dashed, line width=1pt]
\tikzstyle{blocked_edge} = [-{Stealth[length=2mm]}, thick, orange!70, dotted, line width=1.2pt]

\resizebox{0.7\textwidth}{!}{%
\begin{tikzpicture}[node distance=1cm and 1.3cm]

% Shared column headers
\node[above, font=\bfseries] at (1.95, 3.3) {\textbf{Early Training}};
\node[above, font=\bfseries] at (7.15, 3.3) {\textbf{Mid Training}};
\node[above, font=\bfseries] at (12.35, 3.3) {\textbf{Late Training}};

% ============ STOCHASTIC MASKING (TOP ROW) ============
\node[left, font=\large\bfseries] at (-0.8, 1.8) {Global};

% Early training - stochastic
\node[unfrozen_node] (se1a) at (0, 2.8) {};
\node[unfrozen_node] (se1b) at (0, 2) {};
\node[unfrozen_node] (se1c) at (0, 1.2) {};
\node[unfrozen_node] (se1d) at (0, 0.4) {};

\node[unfrozen_node] (se2a) at (1.3, 2.4) {};
\node[frozen_node] (se2b) at (1.3, 1.6) {};
\node[unfrozen_node] (se2c) at (1.3, 0.8) {};

\node[unfrozen_node] (se3a) at (2.6, 2.1) {};
\node[unfrozen_node] (se3b) at (2.6, 1.1) {};

\node[unfrozen_node] (se4a) at (3.9, 2.1) {};
\node[unfrozen_node] (se4b) at (3.9, 1.1) {};

% Edges
\draw[active_edge] (se1a) -- (se2a);
\draw[blocked_edge] (se1a) -- (se2b);
\draw[active_edge] (se1b) -- (se2a);
\draw[blocked_edge] (se1b) -- (se2b);
\draw[active_edge] (se1b) -- (se2c);
\draw[active_edge] (se1c) -- (se2a);
\draw[blocked_edge] (se1c) -- (se2b);
\draw[active_edge] (se1c) -- (se2c);
\draw[frozen_edge] (se1d) -- (se2c);

\draw[active_edge] (se2a) -- (se3a);
\draw[active_edge] (se2a) -- (se3b);
\draw[active_edge] (se2b) -- (se3a);
\draw[active_edge] (se2b) -- (se3b);
\draw[active_edge] (se2c) -- (se3b);

\draw[active_edge] (se3a) -- (se4a);
\draw[active_edge] (se3a) -- (se4b);
\draw[active_edge] (se3b) -- (se4a);
\draw[active_edge] (se3b) -- (se4b);

% Mid training - stochastic
\node[unfrozen_node] (sm1a) at (5.2, 2.8) {};
\node[frozen_node] (sm1b) at (5.2, 2) {};
\node[unfrozen_node] (sm1c) at (5.2, 1.2) {};
\node[unfrozen_node] (sm1d) at (5.2, 0.4) {};

\node[frozen_node] (sm2a) at (6.5, 2.4) {};
\node[unfrozen_node] (sm2b) at (6.5, 1.6) {};
\node[frozen_node] (sm2c) at (6.5, 0.8) {};

\node[frozen_node] (sm3a) at (7.8, 2.1) {};
\node[unfrozen_node] (sm3b) at (7.8, 1.1) {};

\node[unfrozen_node] (sm4a) at (9.1, 2.1) {};
\node[unfrozen_node] (sm4b) at (9.1, 1.1) {};

% Edges
\draw[blocked_edge] (sm1a) -- (sm2a);
\draw[active_edge] (sm1a) -- (sm2b);
\draw[blocked_edge] (sm1b) -- (sm2a);
\draw[frozen_edge] (sm1b) -- (sm2b);
\draw[active_edge] (sm1b) -- (sm2c);
\draw[blocked_edge] (sm1c) -- (sm2a);
\draw[active_edge] (sm1c) -- (sm2b);
\draw[blocked_edge] (sm1c) -- (sm2c);
\draw[active_edge] (sm1d) -- (sm2b);
\draw[blocked_edge] (sm1d) -- (sm2c);

\draw[blocked_edge] (sm2a) -- (sm3a);
\draw[active_edge] (sm2a) -- (sm3b);
\draw[active_edge] (sm2b) -- (sm3a);
\draw[frozen_edge] (sm2b) -- (sm3b);
\draw[frozen_edge] (sm2c) -- (sm3a);
\draw[active_edge] (sm2c) -- (sm3b);

\draw[active_edge] (sm3a) -- (sm4a);
\draw[active_edge] (sm3a) -- (sm4b);
\draw[active_edge] (sm3b) -- (sm4a);
\draw[active_edge] (sm3b) -- (sm4b);

% Late training - stochastic
\node[frozen_node] (sl1a) at (10.4, 2.8) {};
\node[frozen_node] (sl1b) at (10.4, 2) {};
\node[unfrozen_node] (sl1c) at (10.4, 1.2) {};
\node[frozen_node] (sl1d) at (10.4, 0.4) {};

\node[frozen_node] (sl2a) at (11.7, 2.4) {};
\node[frozen_node] (sl2b) at (11.7, 1.6) {};
\node[unfrozen_node] (sl2c) at (11.7, 0.8) {};

\node[frozen_node] (sl3a) at (13, 2.1) {};
\node[unfrozen_node] (sl3b) at (13, 1.1) {};

\node[frozen_node] (sl4a) at (14.3, 2.1) {};
\node[unfrozen_node] (sl4b) at (14.3, 1.1) {};

% Edges
\draw[frozen_edge] (sl1a) -- (sl2a);
\draw[blocked_edge] (sl1a) -- (sl2b);
\draw[frozen_edge] (sl1b) -- (sl2a);
\draw[frozen_edge] (sl1b) -- (sl2b);
\draw[active_edge] (sl1b) -- (sl2c);
\draw[blocked_edge] (sl1c) -- (sl2a);
\draw[frozen_edge] (sl1c) -- (sl2b);
\draw[frozen_edge] (sl1c) -- (sl2c);
\draw[frozen_edge] (sl1d) -- (sl2a);
\draw[frozen_edge] (sl1d) -- (sl2b);
\draw[active_edge] (sl1d) -- (sl2c);

\draw[blocked_edge] (sl2a) -- (sl3a);
\draw[frozen_edge] (sl2a) -- (sl3b);
\draw[blocked_edge] (sl2b) -- (sl3a);
\draw[active_edge] (sl2b) -- (sl3b);
\draw[active_edge] (sl2c) -- (sl3b);
\draw[frozen_edge] (sl2c) -- (sl3a);

\draw[blocked_edge] (sl3a) -- (sl4a);
\draw[active_edge] (sl3b) -- (sl4a);
\draw[active_edge] (sl3b) -- (sl4b);
\draw[active_edge] (sl3a) -- (sl4b);

% ============ LAYER-WISE FORWARD MASKING (BOTTOM ROW) ============
\begin{scope}[yshift=2.5cm]

\node[left, font=\large\bfseries] at (-0.8, -4.2) {Layer-wise};

% Early training - forward
\node[frozen_node] (fe1a) at (0, -3.2) {};
\node[frozen_node] (fe1b) at (0, -4) {};
\node[frozen_node] (fe1c) at (0, -4.8) {};
\node[frozen_node] (fe1d) at (0, -5.6) {};

\node[unfrozen_node] (fe2a) at (1.3, -3.6) {};
\node[unfrozen_node] (fe2b) at (1.3, -4.4) {};
\node[unfrozen_node] (fe2c) at (1.3, -5.2) {};

\node[unfrozen_node] (fe3a) at (2.6, -3.9) {};
\node[unfrozen_node] (fe3b) at (2.6, -4.9) {};

\node[unfrozen_node] (fe4a) at (3.9, -3.9) {};
\node[unfrozen_node] (fe4b) at (3.9, -4.9) {};

% Edges
\draw[active_edge] (fe1a) -- (fe2a);
\draw[active_edge] (fe1a) -- (fe2b);
\draw[active_edge] (fe1b) -- (fe2a);
\draw[active_edge] (fe1b) -- (fe2b);
\draw[active_edge] (fe1b) -- (fe2c);
\draw[active_edge] (fe1c) -- (fe2b);
\draw[active_edge] (fe1c) -- (fe2c);
\draw[active_edge] (fe1d) -- (fe2b);
\draw[active_edge] (fe1d) -- (fe2c);

\draw[active_edge] (fe2a) -- (fe3a);
\draw[active_edge] (fe2a) -- (fe3b);
\draw[active_edge] (fe2b) -- (fe3a);
\draw[active_edge] (fe2b) -- (fe3b);
\draw[active_edge] (fe2c) -- (fe3b);

\draw[active_edge] (fe3a) -- (fe4a);
\draw[active_edge] (fe3a) -- (fe4b);
\draw[active_edge] (fe3b) -- (fe4a);
\draw[active_edge] (fe3b) -- (fe4b);

% Mid training - forward
\node[frozen_node] (fm1a) at (5.2, -3.2) {};
\node[frozen_node] (fm1b) at (5.2, -4) {};
\node[frozen_node] (fm1c) at (5.2, -4.8) {};
\node[frozen_node] (fm1d) at (5.2, -5.6) {};

\node[frozen_node] (fm2a) at (6.5, -3.6) {};
\node[frozen_node] (fm2b) at (6.5, -4.4) {};
\node[frozen_node] (fm2c) at (6.5, -5.2) {};

\node[unfrozen_node] (fm3a) at (7.8, -3.9) {};
\node[unfrozen_node] (fm3b) at (7.8, -4.9) {};

\node[unfrozen_node] (fm4a) at (9.1, -3.9) {};
\node[unfrozen_node] (fm4b) at (9.1, -4.9) {};

% Edges
\draw[frozen_edge] (fm1a) -- (fm2a);
\draw[frozen_edge] (fm1a) -- (fm2b);
\draw[frozen_edge] (fm1b) -- (fm2a);
\draw[frozen_edge] (fm1b) -- (fm2b);
\draw[frozen_edge] (fm1b) -- (fm2c);
\draw[frozen_edge] (fm1c) -- (fm2b);
\draw[frozen_edge] (fm1c) -- (fm2c);
\draw[frozen_edge] (fm1d) -- (fm2b);
\draw[frozen_edge] (fm1d) -- (fm2c);

\draw[frozen_edge] (fm2a) -- (fm3a);
\draw[frozen_edge] (fm2a) -- (fm3b);
\draw[frozen_edge] (fm2b) -- (fm3a);
\draw[frozen_edge] (fm2b) -- (fm3b);
\draw[frozen_edge] (fm2c) -- (fm3b);

\draw[active_edge] (fm3a) -- (fm4a);
\draw[active_edge] (fm3a) -- (fm4b);
\draw[active_edge] (fm3b) -- (fm4a);
\draw[active_edge] (fm3b) -- (fm4b);

% Late training - forward
\node[frozen_node] (fl1a) at (10.4, -3.2) {};
\node[frozen_node] (fl1b) at (10.4, -4) {};
\node[frozen_node] (fl1c) at (10.4, -4.8) {};
\node[frozen_node] (fl1d) at (10.4, -5.6) {};

\node[frozen_node] (fl2a) at (11.7, -3.6) {};
\node[frozen_node] (fl2b) at (11.7, -4.4) {};
\node[frozen_node] (fl2c) at (11.7, -5.2) {};

\node[frozen_node] (fl3a) at (13, -3.9) {};
\node[frozen_node] (fl3b) at (13, -4.9) {};

\node[unfrozen_node] (fl4a) at (14.3, -3.9) {};
\node[unfrozen_node] (fl4b) at (14.3, -4.9) {};

% Edges
\draw[frozen_edge] (fl1a) -- (fl2a);
\draw[frozen_edge] (fl1a) -- (fl2b);
\draw[frozen_edge] (fl1b) -- (fl2a);
\draw[frozen_edge] (fl1b) -- (fl2b);
\draw[frozen_edge] (fl1b) -- (fl2c);
\draw[frozen_edge] (fl1c) -- (fl2b);
\draw[frozen_edge] (fl1c) -- (fl2c);
\draw[frozen_edge] (fl1d) -- (fl2b);
\draw[frozen_edge] (fl1d) -- (fl2c);

\draw[frozen_edge] (fl2a) -- (fl3a);
\draw[frozen_edge] (fl2a) -- (fl3b);
\draw[frozen_edge] (fl2b) -- (fl3a);
\draw[frozen_edge] (fl2b) -- (fl3b);
\draw[frozen_edge] (fl2c) -- (fl3b);

\draw[frozen_edge] (fl3a) -- (fl4a);
\draw[frozen_edge] (fl3a) -- (fl4b);
\draw[frozen_edge] (fl3b) -- (fl4a);
\draw[frozen_edge] (fl3b) -- (fl4b);
\end{scope}

% ============ LEGEND ============
\begin{scope}[shift={(18, 0.5)}]  % Moved up 0.5 units
    % Legend box with rounded corners (no title needed)
    \node[draw, rectangle, rounded corners=5pt, thick, minimum width=3.2cm, minimum height=6.5cm, 
          fill=white, align=left] at (0, -0.5) {};
    
    % Unfrozen node
    \node[unfrozen_node] (leg_unfroz) at (-0.8, 2) {};
    \node[right, align=left] at (-0.3, 2) {Unfrozen\\node};
    
    % Frozen node
    \node[frozen_node] (leg_froz) at (-0.8, 0.8) {};
    \node[right, align=left] at (-0.3, 0.8) {Frozen\\node};
    
    % Active edge
    \draw[active_edge] (-1.2, -0.4) -- (-0.4, -0.4);
    \node[right, align=left] at (-0.3, -0.4) {Active\\gradient};
    
    % Frozen edge
    \draw[frozen_edge] (-1.2, -1.6) -- (-0.4, -1.6);
    \node[right, align=left] at (-0.3, -1.6) {Frozen\\gradient};
    
    % Blocked edge
    \draw[blocked_edge] (-1.2, -2.8) -- (-0.4, -2.8);
    \node[right, align=left] at (-0.3, -2.8) {Blocked\\gradient};
\end{scope}

\end{tikzpicture}
}

\caption{Comparison of masking strategies in StoMPP. \textbf{Top:} Global stochastic masking randomly freezes activations (nodes) and weights (edges) throughout training. \textbf{Bottom:} Layer-wise stochastic masking freezes entire layers sequentially from input to output. Blue indicates active gradient paths, red indicates frozen elements (sign function), and orange indicates edges blocked by frozen target nodes.}
\label{fig:masking_comparison}
\end{figure*}

Motivated by the STE mismatch, we investigate whether \emph{progressive freezing} can serve as an alternative training paradigm for binary networks trained \emph{from scratch}. Progressive and soft-to-hard methods (e.g., incremental quantization, proximal objectives, and annealing schedules) offer principled ways to transition from continuous to discrete representations \cite{INQ_Zhou_2017a, AlphaBlend_Liu_Mattina_2019, Bai_Wang_Liberty_2019a, Yin_Zhang_Lyu_Osher_Qi_Xin_2018, Soft_then_hard_Guo_2024, Self_Binarizing_Lahoud_2019}, but their extension to full BNNs at depth is non-trivial. In particular, we find that \emph{global} progressive freezing, even with stochastic masking, can train BWNs, but fails to train full BNNs due to \emph{activation-induced gradient blockades}: once a unit is committed to a hard step function, gradients cannot propagate through it to earlier layers, preventing learning in deep networks (Fig. \ref{fig:masking_comparison}).

We introduce StoMPP (\textbf{Sto}chastic \textbf{M}asked \textbf{P}artial \textbf{P}rogressive Binarization), a layerwise progressive freezing procedure with stochastic masking. StoMPP progressively \emph{hardens} both weights and activations from a differentiable clipped operator to a hard binary operator, while backpropagating only through the unfrozen (clipped) subset. In contrast to STE training, StoMPP does not apply a hard binary operator in the forward pass and a different surrogate derivative in the backward pass; instead, gradients correspond to the forward operator used on trainable components, and committed components are treated as constants. This yields an estimator-free progressive training paradigm for both BWNs and full BNNs, especially beneficial at greater depth.

Our key contributions are listed below. 

\begin{enumerate}[nosep] 
  \item We identify \textbf{activation-induced gradient blockades} as the key failure mode of global progressive freezing in full BNNs, while BWNs remain trainable under global schedules.
  \item We propose \textbf{layerwise} progressive freezing with stochastic masking, mitigating gradient blockades by scheduling commitment from input to output.
  \item Under a matched \textbf{minimal} training recipe that isolates algorithmic effects, StoMPP improves top-1 accuracy over a BinaryConnect-style STE baseline with gains that increase with depth (ResNet-18: +3.1, +4.7, +1.3 accuracy; ResNet-50: +18.0, +13.5, +3.8 on CIFAR-10/100/ImageNet).
  \item We characterize StoMPP's optimization behavior, including non-monotonic convergence and sensitivity to the freezing schedule.
\end{enumerate}

Under the same minimal recipe (full precision defaults without weight decay or learning rate schedulers), StoMPP achieves 80.9\% top-1 on CIFAR-10 and 53.8\% top-1 on CIFAR-100 with ResNet-18 BNNs, and the improvements over STE are consistent and larger for deeper architectures. We further show StoMPP pairs well with architectural improvements, applying it to Bi-Real Net and suggesting progressive freezing is largely orthogonal to model design.
\section{Related Work}

\textbf{Binary Neural Networks and STE.}
Binary neural networks (BNNs) quantize weights and activations to $\{-1,+1\}$, while binary-weight networks (BWNs) quantize only weights. The dominant approach for training BNNs is the straight-through estimator (STE) \cite{ste_bengio_2013}, which applies the non-differentiable $\mathrm{sign}(\cdot)$ in the forward pass and uses a surrogate gradient in the backward pass. STE enabled a wide range of successful BNNs \cite{XNOR_Net_Rastegari_2016, BiReal_Liu_2018, ReActNet_Liu_2020, OvSW_Xiang}, but it introduces an inherent forward/backward mismatch: the backward update does not correspond to the gradient of the discrete forward computation. Empirically, this mismatch is associated with optimization instability and degraded scaling in deeper networks under standard training recipes \cite{BNN_Survey_2020}. Largely, literature improves STE training by proposing better surrogate gradients or gradient shaping to reduce the mismatch and stabilize training \cite{ReCU_Xu_2021, OvSW_Xiang, BiPer_Vargas_2024a, BiReal_Liu_2018, IR_Net_2020}, and stochastic variants such as probabilistic binarization have also been explored \cite{BinaryConnect_Courbariaux_2016}. These methods typically retain surrogate-gradient backpropagation through binarized operations.

\textbf{Differentiable Relaxations and Annealing.}
To avoid STE, several works optimize differentiable relaxations that are annealed toward a hard quantizer \cite{Self_Binarizing_Lahoud_2019, Yin_Zhang_Lyu_Osher_Qi_Xin_2018, Soft_then_hard_Guo_2024}. During the relaxed phase, forward and backward computations are aligned, but training can become sensitive as the relaxation hardens (e.g., brittle schedules or vanishing/exploding gradients near the discrete limit). In contrast, our goal is to train the discrete model without surrogate gradients by explicitly controlling when and where discrete variables are enforced.

\textbf{Progressive Quantization and Freezing.}
Progressive quantization methods convert a continuous model to a quantized one in stages by freezing subsets of parameters and retraining the remainder. INQ \cite{INQ_Zhou_2017a} progressively fixes groups of weights to quantized values in groups deterministically so remaining weights can adapt, and related stochastic methods probabilistically freeze weights during training \cite{Stochastic_Quantization_Dong_2017}. These approaches are largely developed for weight-only quantization and do not address the additional difficulty of \emph{binary activations}, which can block gradient flow in the absence of STE. Alpha-blending
StoMPP extends progressive freezing to full BNNs (weights and activations) without STE by combining (i) a layerwise binarization schedule and (ii) stochastic masking that refreshes a controlled fraction of discrete variables each update, preserving learning signal while maintaining stability. Relatedly, BiTAT \cite{BiTAT_Park_2022} performs layerwise quantization in a \emph{pretrained finetuning} setting and relies on STE, whereas StoMPP targets end-to-end BNN training without STE.

\textbf{Broader Quantization Training Perspectives.}
Beyond BNNs, quantization-aware training (QAT) methods often target multi-bit quantization using differentiable approximations or learned quantizer parameters (e.g., learned step sizes or clipped activations) \cite{DoReFa_Zhou_2018, LSQ_Esser_2020, PACT_Choi_2018, LQ_Net_Zhang_2018}. Quantization is also studied through constrained/discrete optimization lenses, including alternating minimization and ADMM/proximal-style methods \cite{ADMM_NN_Ren_2018, Bai_Wang_Liberty_2019a}. While complementary, these lines typically do not address end-to-end training with \emph{binary activations} in the \textit{absence} of STE, which is the setting we study.

\textbf{Orthogonal Architectural Improvements.}
A complementary direction improves BNN accuracy via architecture changes, such as learned scaling factors \cite{XNOR_Net_Rastegari_2016}, specialized activation designs \cite{ReActNet_Liu_2020}, or modified residual connections and information pathways \cite{IR_Net_2020, BiReal_Liu_2018}. Many of these introduce additional parameters or computation to compensate for binarization \cite{XNOR_Net_Rastegari_2016, XNOR_Plusplus_Bulat_2019, ReActNet_Liu_2020}. StoMPP is orthogonal: we focus on the training procedure for discrete weights and activations without surrogate gradients, and can be combined with such architectural improvements.

Appendix~\ref{sec:app_related_work} presents further related works.
\section{Method}
\label{sec:method}
We introduce \textbf{StoMPP} (\textbf{Sto}chastic \textbf{M}asked \textbf{P}artial \textbf{P}rogressive binarization), an STE-free BNN training procedure that progressively enforces binarization while preserving learning via \textbf{(i)} stochastic masked partial freezing (soft refresh) and \textbf{(ii)} layerwise scheduling to avoid activation-induced gradient blockades.

\subsection{StoMPP Overview}
We first define StoMPP's masked binarization and its non-STE gradient flow, which are used by both stochastic masked freezing and layerwise scheduling.

\textbf{Forward Mapping.}
StoMPP maintains underlying real-valued variables and uses a mask to decide whether each entry is treated as discrete (frozen) or continuous (unfrozen). Let $u$ denote either a weight entry or a pre-activation entry. We define the continuous proxy
$\mathrm{clip}(u)=\max(-1,\min(1,u))$
and the binary map $\mathrm{sign}(u)\in\{-1,+1\}$.
Given a binary mask $M\in\{0,1\}$ (same shape as $u$), the forward value is
\begin{equation}
\label{eq:stompp_forward}
u' \;=\; M \odot \mathrm{sign}(u) \;+\; (1-M)\odot \mathrm{SmoothFunc}(u),
\end{equation}
where $\odot$ is elementwise multiplication.
We use $\mathrm{clip}$ and $\mathrm{identity}$ as $\mathrm{SmoothFunc}$ for activations and weights, respectively, with $\mathrm{sign}$ in the forward pass to match canonical STE-style BNN parameterizations. StoMPP is independent to these choices and can pair with alternative binarizers (e.g., \cite{BiReal_Liu_2018}).

\textbf{Backward Pass without STE.}
StoMPP does \emph{not} use a surrogate gradient for $\mathrm{sign}(\cdot)$. Instead, it backpropagates only through the continuous proxy branch in Eq.~\eqref{eq:stompp_forward}; frozen entries receive zero gradient (since $\frac{\partial\,\mathrm{sign}(u)}{\partial u}=0$ almost everywhere). Concretely, for activations,

\begin{equation}
\label{eq:stompp_backward}
\frac{\partial \mathcal{L}}{\partial u}
\;=\;
(1-M)\odot \frac{\partial \mathcal{L}}{\partial u'} \odot \frac{\partial\,\mathrm{clip}(u)}{\partial u}.
\end{equation}
Thus StoMPP computes the exact gradient of the proxy mapping for unfrozen entries and assigns zero gradient to frozen entries; it does \emph{not} claim to compute gradients of the fully discrete network. For weights, we apply the same process with the $\mathrm{identity}$ function in place of $\mathrm{clip}$.

\subsection{Stochastic Masked Progressive Freezing}
This component progressively increases the fraction of frozen (binary) entries within a layer while softly refreshing which specific entries are frozen, balancing stability and exploration during training.

\textbf{Masked Variables.}
For each scheduled layer $i$, StoMPP maintains either a weight mask $M_i^{W}$ with the same shape as weights $W_i$ for a weight layer, or an activation mask $M_i^{A}$ with the same shape as the layer pre-activations $z_i$ for an activation layer. We apply StoMPP sequentially to both weights and activations of scheduled layers (unless otherwise stated). For a weight tensor of size $n\times m$, $M_i^W$ has the same shape; for an activation vector of size $n$, $M_i^A$ is size $n$ (and analogously for convolutional tensors). Unless stated otherwise, masking is applied \emph{elementwise}.

\textbf{Freezing Schedule.}
Within a layer's transition, StoMPP targets an increasing frozen fraction $p(\tau)\in[0,1]$ over transition step $\tau=1,\ldots,T$. We use a cubic schedule ${p(\tau)=\left(\frac{\tau}{T}\right)^3}$
% \begin{equation}
%p(\tau)=\left(\frac{\tau}{T}\right)^3,
% \end{equation}
and recommend any monotonically increasing schedule from $0$ to $1$ ending in a fully binary layer ($p(T)=1$).

\textbf{Soft Refresh.}
To prevent premature commitment to a particular frozen configuration, StoMPP \emph{soft-refreshes} the mask each step. For a tensor with $n$ entries, we resample only $k=\lfloor n/r\rfloor$ randomly chosen indices, redrawing those mask values from $\mathrm{Bernoulli}(p(\tau))$ while keeping all other indices unchanged; this preserves the target frozen fraction \emph{in expectation} and yields temporal stability (e.g., $r=100$ updates $\approx 1\%$ of entries per step). This differs from full resampling, where the entire mask changes each step, and from deterministic freezing (e.g., INQ-style progressive quantization), where frozen parameters never change. A slower refresh rate allows the model to adapt to a mostly stable frozen pattern before it is perturbed; we use $r=100$ unless noted otherwise. Section~\ref{sec:hp_ablations} studies the effect of $p(\tau)$ and $r$ and identifies a performance sweet spot balancing stability and exploration.
Algorithm~\ref{algo:layer_masking} summarizes the layer masking procedure, including the soft-refresh update and the resulting forward/backward computation.

\begin{algorithm}[t]
\caption{Layer Masking}
\begin{algorithmic}[1]
\label{algo:layer_masking}
\REQUIRE Freezing schedule $p(t)$, refresh rate $r$
\STATE Initialize masks $M \leftarrow 0$
\FOR{step $t = 1, \ldots, T$}
    \STATE $p \leftarrow p(t)$
    \FOR{each layer}
        \STATE \COMMENT{Soft refresh: update only 1/r fraction of mask}
        \STATE $n \leftarrow$ total parameters in layer
        \STATE $k \leftarrow \lfloor n/r \rfloor$ \COMMENT{Number to resample}
        \STATE Sample $k$ random indices $\mathcal{I} \subset [1,n]$
        \FOR{$i \in \mathcal{I}$}
            \STATE $M[i] \sim \text{Bernoulli}(p)$
        \ENDFOR
        \STATE \COMMENT{Remaining $(1-1/r)$ fraction of $M$ unchanged}
    \ENDFOR
    \STATE
    \STATE \COMMENT{Forward pass}
    \STATE $x' \leftarrow M \odot \text{sign}(x) + (1-M) \odot \text{SmoothFunc}(x)$
    \STATE \COMMENT{Backward pass}
    \STATE $\nabla_x \leftarrow (1-M) \odot \nabla_{x'} \odot \frac{\partial \text{SmoothFunc}}{\partial x}$
    \STATE $x \leftarrow x - \eta \nabla_x$
\ENDFOR
\end{algorithmic}
\end{algorithm}

% \subsection{Layerwise Scheduling to Prevent Gradient Blockades}
\subsection{Layerwise Scheduling for Stable Gradients}
We first describe why globally freezing binary activations can obstruct learning in BNNs, and then present a layerwise schedule that avoids this failure mode.

\textbf{Gradient Blockade under Global Masking.}
In BNNs, freezing an activation uses $a=\mathrm{sign}(z)$, whose derivative is zero almost everywhere. If activations are frozen at arbitrary depths (global masking), gradient signal to earlier layers can be severely attenuated or eliminated along many paths, creating a \textbf{gradient blockade}. This issue is specific to BNNs: in binary weight networks where activations remain continuous (e.g., $\mathrm{clip}$), gradients can still propagate through activation layers. In particular, for our unfrozen continuous activation proxy, $\mathrm{clip(z)}$, is non-zero for most unsaturated activations.
Consider a binary activation layer with output $A = \text{sign}(z)$ where $z$ is the pre-activation. Since $\frac{\partial \text{sign}(z)}{\partial z} = 0$ almost everywhere, 
residual/skip connections may provide alternate routes, but scattered frozen activations can still substantially reduce usable gradient signal in practice.
Fig.~\ref{fig:masking_comparison} illustrates this difference. With global masking (top), frozen activations may appear at arbitrary depths, and the first frozen activation on a path can eliminate gradient signal to earlier layers on that path. This motivates controlling \emph{where} binarization is applied over time.

\textbf{Layerwise Schedule.}
StoMPP avoids blockades by binarizing layers sequentially from input to output. At any time, the network is partitioned into: (1) \textbf{Frozen prefix} ($1,\ldots,\ell-1$): fully binarized %($M_i^{W}=M_i^{A}=\mathbf{1}$)
($M_i=\mathbf{1}$)
; gradients to these layers are not required; (2) \textbf{Transition layer} ($\ell$): partially frozen, updated by SoftRefresh toward $p(\tau)$; and (3) \textbf{Unfrozen suffix} ($\ell+1,\ldots,N$): fully continuous %($M_i^{W}=M_i^{A}=\mathbf{0}$), 
($M_i=\mathbf{0}$)
providing a gradient path from the loss to layer $\ell$.
% \end{itemize}
This ensures the transitioning layer always receives a valid learning signal through its unfrozen entries (Eq.\eqref{eq:stompp_backward}), while frozen layers avoid backpropagating through binary activations. Fig.~\ref{fig:masking_comparison} (bottom) illustrates this: binarization advances as a contiguous input-to-output wave, so unfrozen downstream layers preserve a gradient path for the layer currently transitioning.

\begin{algorithm}[t]
\caption{StoMPP}
\label{algo:stompp}
\begin{algorithmic}[1]
\REQUIRE Layers $L_1, \ldots, L_N$, epochs per layer $E$, schedule $p(t)$, refresh rate $r$
\STATE Initialize all activation masks $M_i \leftarrow 0$ 
\FOR{epoch $e = 1, \ldots, N \times E$}
    \STATE $\ell \leftarrow \lfloor e / E \rfloor$ \COMMENT{Currently quantizing layer $\ell$}
    \FOR{training step $t$ in epoch $e$}
        \FOR{activation layer $i = 1, \ldots, N$}
            \IF{$i < \ell$}
                \STATE $M_i \leftarrow 1$ \COMMENT{Fully frozen (binary)}
            \ELSIF{$i = \ell$}
                \STATE Update $M_i$ using Layer Masking (Algorithm \ref{algo:layer_masking}) \COMMENT{Transitioning}
            \ELSE
                \STATE $M_i \leftarrow 0$ \COMMENT{Not started (continuous)}
            \ENDIF
        \ENDFOR
        \STATE Forward/backward as in Algorithm 1
    \ENDFOR
\ENDFOR
\end{algorithmic}
\end{algorithm}
Algorithm~\ref{algo:stompp} gives the complete StoMPP procedure with layerwise scheduling, applying Algorithm~\ref{algo:layer_masking} sequentially to each layer. For simplicity, we allocate an equal number of epochs/steps to each layer's transition; exploring non-uniform schedules is left to future work.

\subsection{\textbf{Practical Considerations and Overhead.}}
StoMPP is an architecture-agnostic training method. We summarize practical choices for applying it to modern feedforward networks, including which layers to schedule, scheduling granularity, and computational overhead. StoMPP adds minimal overhead: SoftRefresh samples $O(n/r)$ indices per step, and Eq.\eqref{eq:stompp_forward} is elementwise. By the end of training, all scheduled layers satisfy $p(T)=1$, yielding a fully binarized inference network (weights and activations use $\mathrm{sign}$ throughout scheduled layers). Default hyperparameters are provided in Appendix \ref{sec:appendix_a}.

\section{Experiments}
\label{sec:experiments}

\subsection{Experimental Setup}
\label{sec:exp_setup}

We evaluate StoMPP against standard STE under a controlled protocol using ResNets of varying depth (R18/R34/R50) on CIFAR-10, CIFAR-100 \cite{cifar10_krizhevsky}, and ImageNet \cite{imagenet_russakovsky2015}. Unless stated otherwise, all methods share the same backbone, data preprocessing/augmentation, optimizer, batch size, training length, and evaluation procedure, with no added scaling factors, regularization, or specialized activations. The only method-specific difference is the binarization/training rule (StoMPP vs. STE). We also evaluate StoMPP on Bi-Real Net to test architectural compatibility; these results are reported separately in Section~\ref{sec:bireal_modern}.

\noindent\textbf{Networks and Quantization Settings.}
We consider two settings: binary weight networks (BWN; weights binarized, activations real-valued) and binary neural networks (BNN; both weights and activations binarized). Following common practice in the BNN literature, we keep the first and last layers in full precision, and we also keep downsampling/projection layers in full precision, since these layers are empirically more sensitive to quantization and can cause disproportionate accuracy degradation when binarized. All remaining layers follow the method-specific binarization rule. We report Top-1 accuracy on the quantized network.

\noindent\textbf{Baselines.}
Our primary baseline is BinaryConnect or BinaryNet-style STE training with an vanilla ResNet architecture: weights and/or activations are binarized in the forward pass using $\mathrm{sign}(\cdot)$, and gradients are propagated using an STE surrogate (identity) in the backward pass to align with BinaryConnect \cite{BinaryConnect_Courbariaux_2016}. For fairness, the STE baseline uses the same training recipe and precision policy described above. In addition to this main baseline, our ablation study evaluate hybrids that mix StoMPP and STE (Section~\ref{sec:hybrids}) and compatibility with modern BNN components (Section~\ref{sec:bireal_modern}, Bi-Real/OvSW).

\noindent\textbf{Training Recipe and Controlled Deviations from FP.}
For each dataset we start from a standard full-precision ResNet training setup (SGD, lr=0.1, momentum=0.9) and apply it to \emph{all} quantized methods, with two controlled modifications used throughout our binary experiments: (1) \textbf{No weight decay.} In BWNs/BNNs, $\ell_2$ regularization can counteract the desired concentration of weights near $\pm 1$ by pulling parameters toward zero. To avoid introducing a confound that methods may tolerate differently, we disable weight decay for all binary runs (StoMPP and STE).
(2) \textbf{Constant learning rate in the main comparison.} Learning-rate schedules can interact strongly with progressive/scheduled quantization, making it unclear whether improvements come from the binarization method or from schedule tuning. To attribute differences primarily to StoMPP's schedule rather than LR annealing, we keep the learning rate constant for the main StoMPP vs.\ STE comparison. Many scheduled techniques \cite{Self_Binarizing_Lahoud_2019, Stochastic_Quantization_Dong_2017} relies on learning-rate schedules ; we leave more complex schedulers and masking schemes to future work.
% \end{enumerate}

\noindent\textbf{StoMPP Configuration.}
Unless stated otherwise, StoMPP uses a layerwise progression over the quantized portion of the network (input$\rightarrow$output), with stochastic masking and refresh rate $r$ as described in Section~\ref{sec:method}. We use the same schedule family and default hyperparameters across datasets; ablations over schedule shape and refresh rate are reported in Section~\ref{sec:hp_ablations}.
Appendix~\ref{sec:appendix_a} lists per-dataset hyperparameters (e.g., optimizer settings, learning rate, batch size, epochs, and augmentation, and any data preprocessing).

\subsection{Main Results: StoMPP and STE}
\label{sec:main_results}

Table~\ref{tab:main_results} compares StoMPP with a fully-specified STE baseline on BWNs and BNNs across CIFAR-10, CIFAR-100, and ImageNet for ResNet18/34/50. Overall, StoMPP matches or improves over STE and exhibits markedly improved depth scaling in the fully-binarized (BNN) regime.

\noindent\textbf{Binary Neural Networks (BNN).}
StoMPP substantially reduces the degradation with depth observed under STE. On CIFAR-10, STE drops from $77.8\%$ (R18) to $51.5\%$ (R50), whereas StoMPP drops from $80.9\%$ to $69.5\%$. On CIFAR-100, STE declines from $49.1\%$ to $26.7\%$, while StoMPP declines from $53.8\%$ to $40.2\%$. On ImageNet, StoMPP improves over STE at depth (R50: $34.2\%$ vs.\ $30.8\%$), indicating that the trend persists beyond CIFAR. We hypothesize that this depth sensitivity is consistent with the difficulty of optimizing binary activations under an STE gradient approximation; we test this hypothesis via ordering and hybrid ablations in Sections~\ref{sec:layerwise_ablation} and~\ref{sec:hybrids}. These runs use a deliberately simple recipe (no distillation or architecture-specific modifications), so our focus is the relative depth trend between StoMPP and STE.

\begin{table}[t]
\centering
\caption{StoMPP vs STE on BWN and BNN across datasets. 
StoMPP (forward layerwise) demonstrates improved depth scaling, particularly for BNN 
where STE degrades significantly. 
All accuracies (\%) are test accuracy on quantized networks.}
\label{tab:main_results}
\resizebox{0.49\textwidth}{!}{%
\begin{tabular}{|lccccccccc|}
\hline
\multicolumn{1}{|l|}{}                & \multicolumn{3}{c|}{\textbf{CIFAR-10}}                                                                       & \multicolumn{3}{c|}{\textbf{CIFAR-100}}                                                                      & \multicolumn{3}{c|}{\textbf{ImageNet}}                                         \\ \hline
\multicolumn{1}{|l|}{\textbf{Method}} & \multicolumn{1}{c|}{R18}           & \multicolumn{1}{c|}{R34}           & \multicolumn{1}{c|}{R50}           & \multicolumn{1}{c|}{R18}           & \multicolumn{1}{c|}{R34}           & \multicolumn{1}{c|}{R50}           & \multicolumn{1}{c|}{R18}           & \multicolumn{1}{c|}{R34}  & R50           \\ \hline
\multicolumn{10}{|c|}{\cellcolor[HTML]{C0C0C0}\textbf{BWN}}                                                                                                                                                                                                                                                                                          \\ \hline
\multicolumn{1}{|l|}{STE}             & \multicolumn{1}{c|}{89.8}          & \multicolumn{1}{c|}{\textbf{89.8}} & \multicolumn{1}{c|}{88.3}          & \multicolumn{1}{c|}{64.6}          & \multicolumn{1}{c|}{64.9}          & \multicolumn{1}{c|}{64.3}          & \multicolumn{1}{c|}{\textbf{65.5}}          & \multicolumn{1}{c|}{---}  & \textbf{67.8}          \\ \hline
\multicolumn{1}{|l|}{StoMPP}          & \multicolumn{1}{c|}{\textbf{90.7}} & \multicolumn{1}{c|}{89.4}          & \multicolumn{1}{c|}{\textbf{91.2}} & \multicolumn{1}{c|}{\textbf{69.5}} & \multicolumn{1}{c|}{\textbf{66.3}} & \multicolumn{1}{c|}{\textbf{69.0}} & \multicolumn{1}{c|}{60.6}   & \multicolumn{1}{c|}{---}  & {67.3}  \\ \hline
\multicolumn{10}{|c|}{\cellcolor[HTML]{C0C0C0}\textbf{BNN}}                                                                                                                                                                                                                                                                                          \\ \hline
\multicolumn{1}{|l|}{STE}             & \multicolumn{1}{c|}{77.8}          & \multicolumn{1}{c|}{61.5}          & \multicolumn{1}{c|}{51.5}          & \multicolumn{1}{c|}{{49.1}}          & \multicolumn{1}{c|}{33.7}          & \multicolumn{1}{c|}{26.7}          & \multicolumn{1}{c|}{\textbf{41.9}}          & \multicolumn{1}{c|}{37.0} & 30.8        \\ \hline
\multicolumn{1}{|l|}{StoMPP}          & \multicolumn{1}{c|}{\textbf{80.9}} & \multicolumn{1}{c|}{\textbf{76.0}} & \multicolumn{1}{c|}{\textbf{69.5}} & \multicolumn{1}{c|}{\textbf{53.8}} & \multicolumn{1}{c|}{\textbf{39.8}} & \multicolumn{1}{c|}{\textbf{40.2}} & \multicolumn{1}{c|}{39.5} & \multicolumn{1}{c|}{\textbf{39.7}} & \textbf{34.2} \\ \hline
\end{tabular}

}
\vspace{-1em}
\end{table}
\begin{table}[t]
\centering
\caption{\textbf{Masking Ordering and Policy for BWNs and BNNs} 
Backward layerwise causes BNNs failure, with performance degradation at depth. This isolates binary activations as the source of gradient blockades. Stochastic masking outperforms deterministic for BWN at depth. All models trained on CIFAR-100 for 200 epochs. 
$\dagger$ indicates catastrophic learning collapse, subscript indicates train/test gap.
}
\label{tab:stompp_masking_ablation}
\resizebox{0.49\textwidth}{!}{%
\begin{tabular}{|lccccc|}
\hline
\multicolumn{1}{|l|}{}                                                                              & \multicolumn{1}{c|}{}                                  & \multicolumn{2}{c|}{\textbf{ResNet18}}                                    & \multicolumn{2}{c|}{\textbf{ResNet50}}              \\ \cline{3-6} 
\multicolumn{1}{|l|}{\multirow{-2}{*}{\textbf{Schedule}}}                                           & \multicolumn{1}{c|}{\multirow{-2}{*}{\textbf{Policy}}} & \multicolumn{1}{c|}{\textbf{Train}} & \multicolumn{1}{c|}{\textbf{Test}}  & \multicolumn{1}{c|}{\textbf{Train}} & \textbf{Test} \\ \hline
\multicolumn{6}{|c|}{\cellcolor[HTML]{C0C0C0}\textbf{BNN}}                                                                                                                                                                                                                                     \\ \hline
\multicolumn{1}{|l|}{Global}                                                                        & \multicolumn{1}{c|}{Stochastic}                        & \multicolumn{1}{c|}{62.3}           & \multicolumn{1}{c|}{53.2$_{9.1}$}           & \multicolumn{1}{c|}{39.5}           & 35.9$_{3.6}$          \\ \hline
\multicolumn{1}{|l|}{Layerwise}                                                                     & \multicolumn{1}{c|}{Stochastic}                        & \multicolumn{1}{c|}{83.3}           & \multicolumn{1}{c|}{\textbf{53.8}$_{29.5}$}           & \multicolumn{1}{c|}{54.6}           & \textbf{40.0}$_{14.6}$          \\ \hline
\multicolumn{1}{|l|}{\begin{tabular}[c]{@{}l@{}}Reverse \\ Layerwise\end{tabular}}                  & \multicolumn{1}{c|}{Stochastic}                        & \multicolumn{1}{c|}{28.4$^\dagger$} & \multicolumn{1}{c|}{28.4$^\dagger$$_{0.0}$} & \multicolumn{1}{c|}{9.6$^\dagger$}  & 8.6$^\dagger$$_{1.0}$ \\ \hline
\multicolumn{6}{|c|}{\cellcolor[HTML]{C0C0C0}\textbf{BWN}}                                                                                                                                                                                                                                     \\ \hline
\multicolumn{1}{|l|}{\multirow{2}{*}{Global}}                                      & \multicolumn{1}{c|}{Deterministic}                    & \multicolumn{1}{c|}{99.5}           & \multicolumn{1}{c|}{70.2$_{29.3}$}                           & \multicolumn{1}{c|}{96.9}           & 65.9$_{31.0}$                           \\ \cline{2-6} 
\multicolumn{1}{|l|}{}                                                                              & \multicolumn{1}{c|}{Stochastic}                        & \multicolumn{1}{c|}{99.2}           & \multicolumn{1}{c|}{\textbf{70.3}$_{28.9}$}           & \multicolumn{1}{c|}{96.1}           & 67.5$_{28.6}$          \\ \hline
\multicolumn{1}{|l|}{}                                                                              & \multicolumn{1}{c|}{Deterministic}                     & \multicolumn{1}{c|}{99.9}           & \multicolumn{1}{c|}{69.6$_{30.3}$}           & \multicolumn{1}{c|}{99.8}           & 65.5$_{34.3}$          \\ \cline{2-6} 
\multicolumn{1}{|l|}{\multirow{-2}{*}{Layerwise}}                                                   & \multicolumn{1}{c|}{Stochastic}                        & \multicolumn{1}{c|}{99.9}           & \multicolumn{1}{c|}{69.5$_{30.4}$}           & \multicolumn{1}{c|}{99.9}           & \textbf{69.0}$_{30.9}$          \\ \hline
\multicolumn{1}{|l|}{}                                                                              & \multicolumn{1}{c|}{Deterministic}                     & \multicolumn{1}{c|}{99.0}           & \multicolumn{1}{c|}{68.9$_{30.1}$}           & \multicolumn{1}{c|}{99.7}           & 68.4$_{31.3}$          \\ \cline{2-6} 
\multicolumn{1}{|l|}{\multirow{-2}{*}{\begin{tabular}[c]{@{}l@{}}Reverse\\ Layerwise\end{tabular}}} & \multicolumn{1}{c|}{Stochastic}                        & \multicolumn{1}{c|}{99.6}           & \multicolumn{1}{c|}{69.7$_{29.9}$}           & \multicolumn{1}{c|}{99.6}           & 66.6$_{33.0}$          \\ \hline
\end{tabular}
% \begin{tabular}{llc|ccc|ccc}
% \toprule
% \textbf{Network} & \textbf{Schedule} & \textbf{Policy}
% & \multicolumn{3}{c|}{\textbf{ResNet18}}
% & \multicolumn{3}{c}{\textbf{ResNet50}} \\
% & & 
% & \textbf{Train} & \textbf{Test} & \textbf{Gap}
% & \textbf{Train} & \textbf{Test} & \textbf{Gap} \\
% \midrule
% \multirow{3}{*}{\textbf{BNN}}
% & Global & Stochastic
% & 62.3 & 53.2 & 9.1 & 39.5 & 35.9 & 3.6 \\
% & Layerwise & Stochastic
% & 83.3 & 53.8 & 29.5 & 54.6 & 40.0 & 14.6 \\
% & Reverse Layerwise & Stochastic
% & 28.4$^\dagger$ & 28.4$^\dagger$ & 0.0 & 9.6$^\dagger$ & 8.6$^\dagger$ & 1.0 \\
% \midrule
% \multirow{6}{*}{\textbf{BWN}}
% & \multirow{2}{*}{Global}
% & Deterministic
% & 99.5 & 70.2 & 29.3 & 96.9 & 65.9 & 31.0 \\
% & & Stochastic
% & 99.2 & 70.3 & 28.9 & 96.1 & 67.5 & 28.6 \\
% \cmidrule{2-9}
% & \multirow{2}{*}{Layerwise}
% & Deterministic
% & 99.9 & 69.6 & 30.3 & 99.8 & 65.5 & 34.3 \\
% & & Stochastic
% & 99.9 & 69.5 & 30.4 & 99.9 & 69.0 & 30.9 \\
% \cmidrule{2-9}
% & \multirow{2}{*}{Reverse Layerwise}
% & Deterministic
% & 99.0 & 68.9 & 30.1 & 99.7 & 68.4 & 31.3 \\
% & & Stochastic
% & 99.6 & 69.7 & 29.9 & 99.6 & 66.6 & 33.0 \\
% \bottomrule
% \end{tabular}
}
\end{table}
\noindent\textbf{Binary Weight Networks (BWN).}
In the BWN setting (binary weights with real-valued activations), both methods are comparatively stable across depth, and StoMPP matches or improves over STE in most configurations. On CIFAR-10, StoMPP improves at R18 and R50 (R18: $90.7\%$ vs.\ $89.8\%$, R50: $91.2\%$ vs.\ $88.3\%$), with a small deficit at R34. On CIFAR-100, StoMPP improves consistently across depths (R18: $69.5\%$ vs.\ $64.6\%$, R34: $66.3\%$ vs.\ $64.9\%$, R50: $69.0\%$ vs.\ $64.3\%$). We observe a small non-monotonicity at R34 followed by improved performance at R50; since R50 uses bottleneck blocks whereas R18/R34 use basic blocks, architectural differences may contribute, so we focus on overall trends rather than assuming monotonic scaling across these variants.

Overall, the largest gains arise for full BNNs (with binarized activations), where STE exhibits pronounced depth sensitivity. Sections~\ref{sec:layerwise_ablation}, \ref{sec:epoch_ablation}, and \ref{sec:lr_ablation} analyze this behavior through ordering/policy ablations and training dynamics.

\subsection{Ordering Ablation: Layerwise Prevents Collapse}
\label{sec:layerwise_ablation}

Progressive freezing methods often apply a \emph{global} mask (e.g., INQ-style freezing) over all quantized layers. We find that this global masking is effective for BWNs but degrades or fails in BNNs. To isolate the role of progression order in the presence of binary activations, we compare three mask orderings under a fixed training recipe (Table~\ref{tab:stompp_masking_ablation}; Figure~\ref{fig:masking_comparison}).

We evaluate three types of \textit{mask ordering}. (1) \textbf{Layerwise (input$\rightarrow$output):} progressively mask layers from the first quantized layer to the last. This is StoMPP’s default and is used in Section~\ref{sec:main_results}.
(2) \textbf{Global:} apply the same progression schedule to the entire quantized subnetwork at once (INQ-style).
(3) \textbf{Reverse layerwise (output$\rightarrow$input):} progressively mask layers from the last quantized layer to the first, intended to stress-test the effect of blocking gradient flow early in the network.

We use two masking policies: \textbf{(1) Stochastic:} freeze a fraction $p(t)$ at random and refresh a $\tfrac{1}{r}$ subset each step (default StoMPP); \textbf{(2) Deterministic (BWN only):} freeze the $p(t)\%$ of weights closest to $\pm1$ and refresh the frozen set each step. We do not apply (2) to BNNs since activations lack a “closeness to $\pm1$” analogue.

\noindent\textbf{Results and implication.}
In BNNs, the ordering is decisive: \textbf{layerwise} training yields the best performance, \textbf{global} is worse, and \textbf{reverse layerwise} causes catastrophic collapse (Table~\ref{tab:stompp_masking_ablation}). For example on CIFAR-100, reverse layerwise collapses to near-chance performance (R18: $28.4\%$, R50: $8.6\%$), while forward layerwise reaches $53.8\%$ (R18) and $40.0\%$ (R50). In contrast, BWNs are far less sensitive to ordering: both forward and reverse layerwise remain competitive, and deterministic masking can be effective at depth.
These results support the interpretation that \emph{binary activations} make training particularly sensitive to progression order: freezing later layers before earlier ones can severely restrict the effective learning signal reaching upstream quantized layers. The gap widens with depth, consistent with the intuition that blockages introduced late in the network can have a larger downstream impact in deeper models.

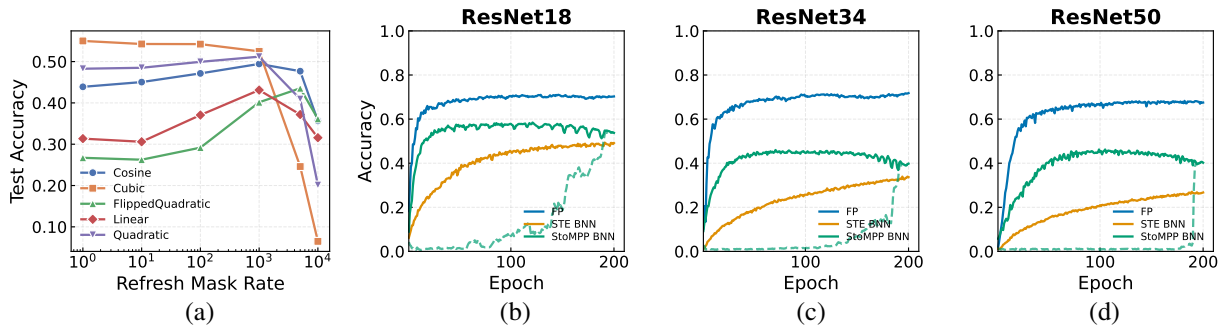
\begin{figure*}[tb]
    \centering
    \begin{overpic}[width=0.95\linewidth]{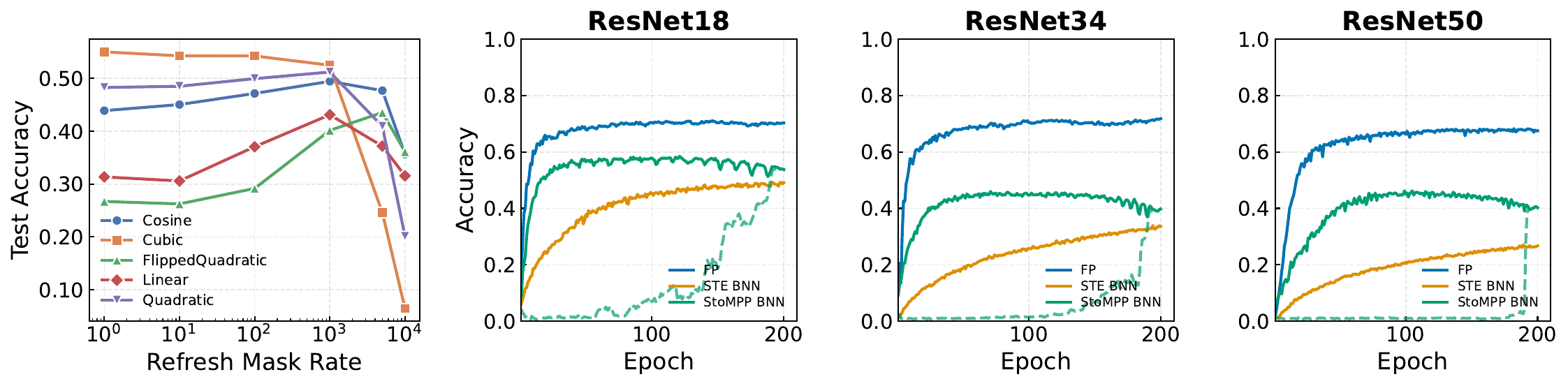}
        \put(15,-1.5){(a)}   % Centered below first subplot
        \put(40.7,-1.5){(b)}   % Centered below second subplot
        \put(64.9,-1.5){(c)}   % Centered below third subplot
        \put(88.8,-1.5){(d)}   % Centered below fourth subplot
    \end{overpic}
    \caption{
        \textbf{(a)} Hyperparameter sweep for BNNs under the \emph{global} mask on CIFAR-100 with ResNet18. We vary the freezing schedule $p(t)$ and refresh rate $r$ for StoMPP and report Top-1 test accuracy (\%).
        \textbf{(b--d)} Accuracy trajectories of CIFAR-100 on ResNets, trained with STE and StoMPP under the same training recipe. StoMPP exhibits a sawtooth pattern corresponding to progressive freezing, while STE improves more smoothly over training. The dashed line represents the fully quantized StoMPP network accuracy.
    }
    \label{fig:hp_sweep_and_curves}
\end{figure*}

\subsection{Hyperparameter Ablation: Schedule \& Refresh}
\label{sec:hp_ablations}
StoMPP introduces a stochastic mask that freezes a fraction of parameters while refreshing a subset throughout training. The mask is governed by (i) a \emph{freezing schedule} $p(t)\in[0,1]$, which specifies the fraction of parameters frozen at step $t$, and (ii) a \emph{refresh rate} $r$, which controls how frequently the frozen set is resampled (we refresh approximately a $1/r$ fraction of masked indices per step). In this section we ablate both $p(t)$ and $r$ to assess sensitivity and to motivate the default settings used in our main experiments.

We evaluate the two masking scopes from Section~\ref{sec:layerwise_ablation}: \emph{layerwise} masking and \emph{global} masking. Since global masking is the more challenging setting for BNNs, we present the primary schedule/refresh sweep for BNNs under the global mask on CIFAR-100 with ResNet18 (Figure~\ref{fig:hp_sweep_and_curves} (b-d)); corresponding layerwise sweeps are reported in Appendix~\ref{sec:app_hp_forward_masking} and show reduced sensitivity.

\noindent\textbf{Schedules.}
Let $T$ denote the total number of schedule steps (global) or the number of steps allocated to a layer (layerwise), and let $t\in\{0,\dots,T\}$ denote the current step. We compare five monotone schedules:\\
(1) \textit{Cosine} $p(t)=\tfrac{1}{2}-\tfrac{1}{2}\cos(\pi t/T)$,\\
(2) \textit{Linear} $p(t)=t/T$,\\
(3) \textit{Quadratic} $p(t)=(t/T)^2$\\
(4) \textit{Cubic} $p(t)=(t/T)^3$, and\\
(5) \textit{Flipped quadratic} $p(t)=2(t/T)-(t/T)^2$.

Figure~\ref{fig:hp_sweep_and_curves} (b-d) shows that performance depends on both schedule shape and refresh rate. Across schedules, accuracy is strongest for moderate refresh rates (roughly $r\in[10^2,10^3]$), while very large refresh rates (e.g., $r=10^4$) substantially degrade performance, indicating that insufficient refreshing can cause training to stall or collapse. Among the tested schedules, the cubic schedule achieves the strongest accuracy over a broad range up to $r\approx 10^3$.
Based on this sweep, we use the \textbf{cubic} schedule as the default and set the refresh rate to \textbf{$r=100$} in our main experiments, which lies in the stable high-performing region of the sweep.

\subsection{Hybrid Ablation: Mixing StoMPP and STE}
\label{sec:hybrids}
StoMPP and STE act on different points in binary training, so we test whether mixing them can improve performance. We consider two hybrids that swap the training rule used for weights and activations: \textbf{Hybrid (A/W)} applies StoMPP to \emph{activations} and STE to \emph{weights}, while \textbf{Reverse Hybrid (W/A)} applies StoMPP to \emph{weights} and STE to \emph{activations}. All methods are trained under the same protocol.

Table~\ref{tab:compare_hybrids} shows that all StoMPP-based variants substantially outperform pure STE across depths. The best choice depends on depth: for R18 and R34, \textbf{Reverse Hybrid (W/A)} attains the highest test accuracy, suggesting that much of the gain comes from improving \emph{weight} optimization while keeping STE for binary activations. For the deeper R50, \textbf{fully StoMPP} achieves the best test accuracy, indicating that as depth increases, applying StoMPP to both weights and activations becomes increasingly important. Applying StoMPP only to activations (Hybrid, A/W) is consistently weaker than the weight-focused variants, reinforcing that weight optimization is the primary driver of gains, while full StoMPP scales most reliably to deeper networks. See Appendix~\ref{sec:appendix_b} for additional observation on different qualitative training-curve behavior in the BWN setting.

% StoMPP and STE modify different parts of the binary training pipeline, so we test whether combining them yields additional gains. We evaluate two hybrids that mix the training rule used for weights and activations: \textbf{Hybrid (A/W)} applies StoMPP to \emph{activations} and STE to \emph{weights}, while \textbf{Reverse Hybrid (W/A)} applies StoMPP to \emph{weights} and STE to \emph{activations}. We compare both hybrids to fully STE and fully StoMPP under the same training protocol. 

% Table~\ref{tab:compare_hybrids} shows that StoMPP and both hybrids outperform pure STE on CIFAR-100 across depths. The strongest results are obtained by applying StoMPP to \emph{weights} while retaining STE for \emph{activations} (Reverse Hybrid, W/A), which achieves the best test accuracy on R18/R34 and remains competitive on R50. In contrast, applying StoMPP only to activations (Hybrid, A/W) underperforms the other StoMPP-based variants, suggesting that in this setup StoMPP contributes most when applied to \emph{weight} optimization while STE remains effective for training binary activations. We additionally observe different qualitative training-curve behavior in the BWN setting (see \rnote{Appendix~\ref{sec:appendix_b}}).

\subsection{Dynamics Ablation: Sawtooth Training Dynamics}
\label{sec:training_dynamics}
Figure~\ref{fig:hp_sweep_and_curves} illustrates that StoMPP and STE exhibit qualitatively different optimization dynamics on CIFAR-100 across network depths. Under STE, test accuracy typically increases smoothly over training, consistent with a fixed binarized forward pass throughout optimization. In contrast, StoMPP produces a characteristic \emph{sawtooth} trajectory: when a new layer enters the freezing phase, accuracy drops, followed by a recovery period as the remaining unfrozen parameters adapt. This pattern repeats as StoMPP progresses through the network until all scheduled layers are binarized.

These dynamics are also reflected in the hybrid ablation (Section~\ref{sec:hybrids}): the reverse hybrid (StoMPP weights, STE activations) exhibits the same sawtooth behavior, whereas the hybrid that applies StoMPP only to activations does not, consistent with StoMPP’s primary effect operating through the progression/freezing of weights. We report additional training curves (including BWN variants) and corresponding training-accuracy trajectories in Appendix~\ref{sec:appendix_b}.

\begin{table}[t]
\centering
\caption{Comparison of hybrid techniques on CIFAR-100.
Entries report \textit{Train / Test} accuracy (\%) on quantized networks. {(A/W) indicates StoMPP activations and STE weights, while (W/A) indicates StoMPP weights and STE activations.}}
\label{tab:compare_hybrids}
\resizebox{0.9\linewidth}{!}{%
\begin{tabular}{l|ccc}
\toprule
\textbf{Method} & R18 & R34 & R50 \\
\midrule
STE
& 66.6 / 49.1 & 37.7 / 33.7 & 28.8 / 26.7 \\
\midrule
StoMPP
& 83.2 / 53.8 & 72.4 / 39.8 & 54.6 / 40.2 \\
Hybrid (A/W)
& 80.5 / 51.8 & 67.6 / 40.9 & 49.3 / 36.4 \\
Reverse Hybrid (W/A)
& 85.3 / 55.8 & 71.6 / 41.0 & 55.6 / 35.6 \\
\bottomrule
\end{tabular}%
}
\vspace{-2em}
\end{table}
\begin{table}[t]
\centering
\caption{\textbf{Learning Rate Sensitivity on CIFAR-100 ResNet-18 BNN.} 
StoMPP is more stable at low learning rates (33.2\% at lr=0.001 vs. STE’s 10.4\%) and reaches a higher peak at lr=0.05 (54.2\% vs. 48.9\%). All models are trained for 200 epochs with a constant learning rate; results report top-1 test accuracy.}
\label{tab:lr_sweep}
\small
\begin{tabular}{@{}ccccc@{}}
\toprule
\multicolumn{1}{l}{Learning Rate} & \multicolumn{1}{l}{0.001} & \multicolumn{1}{l}{0.01} & \multicolumn{1}{l}{0.05} & \multicolumn{1}{l}{0.1} \\ \midrule
STE                               & 10.4                      & 30.9                     & 46.8                     & \textbf{48.9}           \\
StoMPP                            & 33.2                      & 51.5                     & \textbf{54.2}            & 53.8                    \\ \bottomrule
\end{tabular}
\end{table}

\begin{figure}[!htb]
    \centering
    \includegraphics[width=\linewidth]{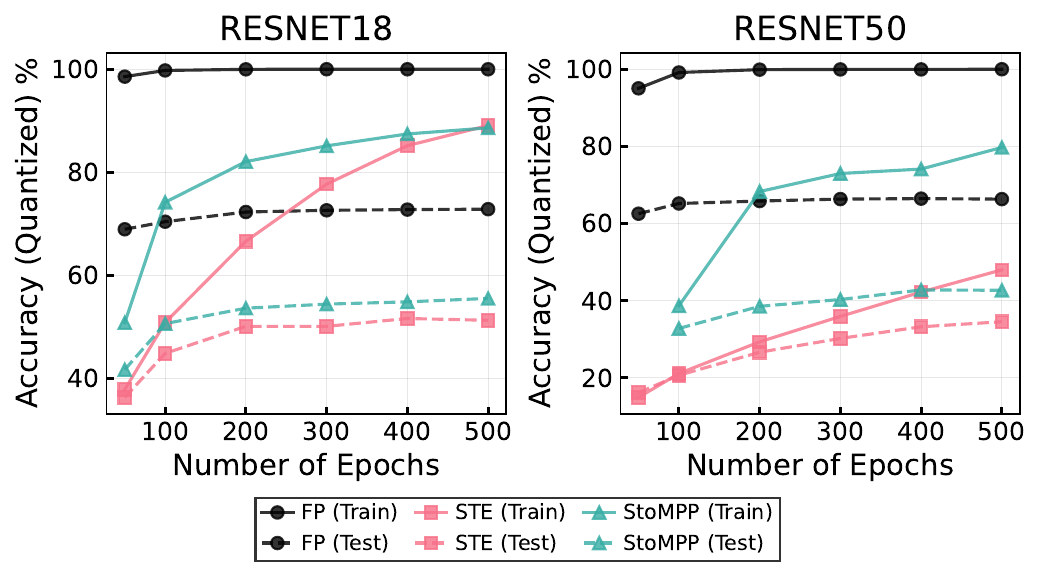}
    \caption{Sweep of epochs under training scheme; solid lines show train accuracy and dashed lines show test accuracy. Note 50 epochs is not included for StoMPP ResNet50 as there are not enough epochs to apply at least one epoch per layer of layerwise masking.}
    \label{fig:epoch_ablation}
    \vspace{-2em}
\end{figure}

\begin{table}[t]
\centering
\caption{Top-1 accuracy (\%) on CIFAR-100 for BiReal18 and BiReal34 under different training schemes and quantization methods. Bi-Real Net primarily reports ImageNet settings; for CIFAR we follow the standard modified ResNet stem used in prior work~\cite{ResNets_He_2015, BiReal_Liu_2018}.} 
\label{tab:birenet-cifar100-all}
\small
\begin{tabular}{lcccc}
\toprule
Training & \multicolumn{2}{c}{BiReal18} & \multicolumn{2}{c}{BiReal34} \\
\cmidrule(lr){2-3} \cmidrule(lr){4-5}
         & StoMPP & STE & StoMPP & STE \\
\midrule
Scratch   & 56.6 & 56.0 & 59.1 & 55.8 \\
Pretrained & 59.4 & 63.5 & 62.3 & 60.6 \\

\bottomrule
\end{tabular}
\vspace{-1em}
\end{table}

\subsection{Epoch Ablation: Accuracy and Training Epochs}
\label{sec:epoch_ablation}

Longer schedules are common for BNNs, so we sweep training epochs and compare STE vs. StoMPP on ResNet18/50, reporting train and test accuracy to separate optimization from generalization. Figure~\ref{fig:epoch_ablation} shows both methods benefit from more epochs, while the full-precision reference converges much faster. StoMPP achieves higher training accuracy early, indicating faster optimization in the low-epoch regime. With sufficient training, STE partially closes the gap on ResNet18 (e.g., by 500 epochs), but on deeper ResNet50 it improves more slowly and remains lower throughout, consistent with stronger depth-induced optimization difficulty. Overall, StoMPP converges in fewer epochs and scales better with depth under the same recipe.

\subsection{Learning-Rate Ablation: Sensitivity Analysis}
\label{sec:lr_ablation}

We ablate the learning rate with all other settings fixed, sweeping $\mathrm{lr}\in{10^{-3},10^{-2},10^{-1}}$ to cover typical BNN/QAT practice~\cite{QNN_Hubara_2016}. We exclude larger rates since they often require extra stabilization (e.g., warmup) that would confound a controlled comparison~\cite{Oscillations_2022}. Table~\ref{tab:lr_sweep} shows StoMPP outperforms STE across learning rates and prefers a slightly smaller optimum, suggesting StoMPP may benefit from tailored recipes.

\subsection{Compatibility with Architectures \& STE Variants}
\label{sec:bireal_modern}
We evaluate whether StoMPP’s gains persist on top of a binary-specific architecture (Bi-Real Net) and how it interact with STE refinements (e.g. OvSW~\cite{OvSW_Xiang}).

We train BiReal18/34 from scratch and finetune with an identical recipe (same optimizer/augmentation; $\mathrm{lr}=0.06$, 300 epochs). Holding the architecture fixed, we vary only the training rule (StoMPP and improved STE; Section~\ref{sec:exp_setup}) under matched compute. Table~\ref{tab:birenet-cifar100-all} shows StoMPP matches or improves over STE, indicating it transfers to binary-specific architectures without additional architectural changes.

We combine StoMPP with OvSW by applying OvSW only to the unfrozen parameters. On CIFAR-10, OvSW improves the STE baseline to 91.3\%, but drops to 9.85\% when paired with StoMPP. This suggests OvSW’s gains are not additive under a naive composition, motivating progressive freezing compatible refinements.

\section{Conclusion}
This work revisits progressive freezing as an alternative to straight-through estimators for training binary networks from scratch, introducing StoMPP, an estimator-free progressive binarization method for both BWNs and full BNNs. StoMPP preserves learning via stochastic masked partial freezing (soft refresh) and avoids failures in full BNNs through an input-to-output layerwise schedule.
Our central finding is that progression order is decisive when activations are binarized: reverse progression collapses on ImageNet ResNet-50 (8.6\%), while forward progression succeeds (40.0\%), implicating activation-induced gradient blockades as the primary failure mode of global or poorly ordered schedules. BWNs show no comparable sensitivity, supporting the conclusion that frozen binary activations are the bottleneck.
Under controlled recipes, StoMPP trains BNNs without surrogate gradients and consistently outperforms STE with gains that grow with depth, while transferring to binary-specific architectures. Future work will evaluate StoMPP under competitive training settings and explore hybrids that use progressive freezing for weights alongside complementary treatments for activations.

\section*{Acknowledgements}

This research utilized computational resources supported
by the Academic \& Research Computing (ARC) group at Worcester Polytechnic Institute.

We'd like to thank Subrata Biswas, Mohammad Nur Hossain Khan, and members of the BASHLab for their feedback, as well as Professor Jacob Whitehill for guidance in the preliminary stages of this research.

\section*{Impact Statement}

This work develops training methods for binary neural networks. It inherits the societal impacts of Machine Learning in general as well as those typical of efficiency improvements in machine learning: enabling deployment on low-power devices with the associated benefits and risks of AI on edge.

\nocite{langley00}

\bibliography{example_paper}
\bibliographystyle{icml2025}

%%%%%%%%%%%%%%%%%%%%%%%%%%%%%%%%%%%%%%%%%%%%%%%%%%%%%%%%%%%%%%%%%%%%%%%%%%%%%%%
%%%%%%%%%%%%%%%%%%%%%%%%%%%%%%%%%%%%%%%%%%%%%%%%%%%%%%%%%%%%%%%%%%%%%%%%%%%%%%%
% APPENDIX
%%%%%%%%%%%%%%%%%%%%%%%%%%%%%%%%%%%%%%%%%%%%%%%%%%%%%%%%%%%%%%%%%%%%%%%%%%%%%%%
%%%%%%%%%%%%%%%%%%%%%%%%%%%%%%%%%%%%%%%%%%%%%%%%%%%%%%%%%%%%%%%%%%%%%%%%%%%%%%%
\newpage
\appendix
\onecolumn

\section{Training Recipes and Configurations}
\label{sec:appendix_a}

As described in the main paper in Section~\ref{sec:exp_setup}, we apply a minimal training condition with minimal modifications (no lr schedule, no weight decay) from a standard full precision schedule. Below are the precise training configurations for each experiments. We use default hyperparameters of $p(t)=(t/T)^3$ and refresh rate $r=100$.

\subsection{Main Results Configuration}
\label{sec:main_results_config}

Table~\ref{tab:training_config_cifar_main} provides the training configurations used for our main results (Table~\ref{tab:main_results}). We report settings for each dataset separately due to differences in resolution, augmentation, and training length. For our CIFAR-10 and CIFAR-100 results, we apply the same training recipe, apart from the dataset.

We apply a similar but marginally different training recipe to ImageNet as shown in Table~\ref{tab:training_config_imagenet_main}. We do this to align with standard full precision practices for ImageNet, again adhering to a minimal training recipe without schedulers or weight decay.

\subsection{Ablation Configuration}
\label{sec:ablation_configuration}
 Table~\ref{tab:experiment_overview} provides the overview of training scheme for the experimental ablations. All of our ablations use the same scheme aside from the controlled variable that is changed (for example, when sweeping epochs in Figure~\ref{fig:epoch_ablation}, this configuration was used only varying the epochs). We outline the modifications each ablation makes to the training scheme in Table~\ref{tab:training_config_ablations}.

\subsection{BiReal-Net Configuration}

For our BiReal-Net, train with a learning rate of $0.06$ over $300$ epochs. These particular values were selected to align with two general prinicples used for training BiReal-Net \cite{BiReal_Liu_2018}.

\begin{enumerate}[nosep]  
\item BiReal-Net uses a lower learning rate than standard ResNets, beginning at 0.01 throughout training.  
\item BiReal-Net is trained for an extended number of epochs when compared to a full precision ImageNet.
\end{enumerate}

For these reasons, we apply many epochs (300 for training from scratch, and an additional 300 for pretraining). We apply this scheme due to the discussed difficulties of applying StoMPP to learning rate schedules (\ref{sec:exp_setup}) and for simplicity. The pretrained networks for both StoMPP and STE use clip activation functions, but apply no freezing and no forward pass quantization, respectively. Aside from these changes ($lr=0.06$, 300 epochs), we use the training recipe described in the Ablation Configuration (\ref{sec:ablation_configuration}).

\begin{table}[H]
\centering
\caption{Training configuration for CIFAR-10 and CIFAR-100 main results (Table~\ref{tab:main_results}). Both datasets use identical settings except for the dataset itself.}
\label{tab:training_config_cifar_main}
\small
\begin{tabular}{ll}
\toprule
\textbf{Category} & \textbf{Setting} \\
\midrule
Dataset & CIFAR-10 / CIFAR-100 \\
Input resolution & $32 \times 32$ \\
Train batch size & 256 \\
Test batch size & 256 \\
Data augmentation & RandomCrop(32, pad=4), HorizontalFlip \\
Normalization & mean (0.5071, 0.4865, 0.4409), std (0.2673, 0.2564, 0.2762) \\
Model architectures & ResNet-18, ResNet-34, ResNet-50 \\
Model variants & STE-BNN, StoMPP-BNN, STE-BWN, StoMPP-BWN \\
Optimizer & SGD (Nesterov momentum) \\
Learning rate & 0.1 \\
Momentum & 0.9 \\
Weight decay & 0 \\
LR schedule & Constant \\
Loss & Cross-entropy \\
Label smoothing & 0.0 \\
Masking scheme (StoMPP) & Layerwise progressive masking \\
Mask schedule (StoMPP) & Cubic ($p: 0 \rightarrow 1$) \\
Mask refresh rate (StoMPP) & 100 \\
Activation function (unfrozen) & Clip \\
Frozen activation & Sign \\
Training epochs & 200 \\
Precision & FP32 \\
\bottomrule
\end{tabular}
\end{table}

\begin{table}[H]
\centering
\caption{Training configuration for ImageNet main results (Table~\ref{tab:main_results}). Training epochs vary by architecture as shown below.}
\label{tab:training_config_imagenet_main}
\small
\begin{tabular}{ll}
\toprule
\textbf{Category} & \textbf{Setting} \\
\midrule
Dataset & ImageNet (ILSVRC2012) \\
Input resolution & $224 \times 224$ \\
Train batch size & 256 \\
Test batch size & 256 \\
Data augmentation & RandomResizedCrop(224), HorizontalFlip \\
Normalization & mean (0.485, 0.456, 0.406), std (0.229, 0.224, 0.225) \\
Model architectures & ResNet-18, ResNet-34, ResNet-50 \\
\textbf{Training epochs} & ResNet-18: 97, ResNet-34: 131, ResNet-50: 98 \\
Model variants & STE-BNN, StoMPP-BNN, STE-BWN, StoMPP-BWN \\
Optimizer & SGD (Nesterov momentum) \\
Learning rate & 0.1 \\
Momentum & 0.9 \\
Weight decay & 0 \\
LR schedule & Constant \\
Loss & Cross-entropy \\
Label smoothing & 0.0 \\
Masking scheme (StoMPP) & Layerwise progressive masking \\
Mask schedule (StoMPP) & Cubic ($p: 0 \rightarrow 1$) \\
Mask refresh rate (StoMPP) & 100 \\
Activation function (unfrozen) & Clip \\
Frozen activation & Sign \\
Precision & FP32 \\
\bottomrule
\end{tabular}
\end{table}

% \begin{table}[h!]
% \centering
% \caption{Dataset-specific training details for main experiments.}
% \label{tab:dataset_training_overrides}
% \small
% \begin{tabular}{llll}
% \toprule
% \textbf{Dataset} & \textbf{Input} & \textbf{Epochs} & \textbf{Data augmentation} \\
% \midrule
% CIFAR-10 &
% $32 \times 32$ &
% 200 &
% RandomCrop(32, pad=4), HorizontalFlip \\

% CIFAR-100 &
% $32 \times 32$ &
% 200 &
% RandomCrop(32, pad=4), HorizontalFlip \\

% ImageNet &
% $224 \times 224$ &
% 90 &
% RandomResizedCrop(224), HorizontalFlip \\
% \bottomrule
% \end{tabular}
% \end{table}

\begin{table}[H]
\centering
\caption{\textbf{Overview of experimental variants and controlled modifications from the base StoMPP scheme in Table~\ref{tab:training_config_ablations}.}
Unless otherwise stated, all experiments use the same backbone, data preprocessing, optimizer, batch size, training protocol, and precision policy described in Section~\ref{sec:exp_setup}. Each variant modifies exactly one component of the base scheme.}
\label{tab:experiment_overview}
\small
\begin{tabular}{llll}
\toprule
\textbf{Experiment} & \textbf{Component} & \textbf{Modification} & \textbf{Reference} \\
\midrule
STE baseline & Training rule & STE instead of StoMPP & Tab.~\ref{tab:main_results} \\

StoMPP (default) & -- & Base scheme (layerwise, cubic, $r{=}100$) & Tab.~\ref{tab:main_results} \\

\midrule
\textbf{Ordering ablation} & Mask ordering &
Layerwise / Global / Reverse & Sec.~\ref{sec:layerwise_ablation} \\

\textbf{Policy ablation} & Mask policy &
Stochastic vs.\ deterministic (BWN only) & Sec.~\ref{sec:layerwise_ablation} \\

\midrule
\textbf{Schedule ablation} & Freezing schedule $p(t)$ &
Cosine / Linear / Quadratic / Cubic / Flipped quad. & Sec.~\ref{sec:hp_ablations} \\

\textbf{Refresh ablation} & Refresh rate $r$ &
$r \in [10, 10^4]$ & Sec.~\ref{sec:hp_ablations} \\

\midrule
\textbf{Hybrid (A/W)} & Training rule split &
StoMPP activations, STE weights & Sec.~\ref{sec:hybrids} \\

\textbf{Hybrid (W/A)} & Training rule split &
StoMPP weights, STE activations & Sec.~\ref{sec:hybrids} \\

\midrule
\textbf{Epoch ablation} & Training length &
Vary total epochs & Sec.~\ref{sec:epoch_ablation} \\

\textbf{LR ablation} & Learning rate &
$\mathrm{lr}\in\{10^{-3},10^{-2},10^{-1}\}$ & Sec.~\ref{sec:lr_ablation} \\

\bottomrule
\end{tabular}
\end{table}

\begin{table}[H]
\centering
\caption{Training configuration for ablations.}
\label{tab:training_config_ablations}
\small
\begin{tabular}{ll}
\toprule
\textbf{Category} & \textbf{Setting} \\
\midrule
Dataset & CIFAR-100 \\
Input resolution & $32 \times 32$ \\
Train batch size & 256 \\
Test batch size & 256 \\
Data augmentation & RandomCrop(32, pad=4), HorizontalFlip \\
Normalization & mean (0.5071, 0.4865, 0.4409), std (0.2673, 0.2564, 0.2762) \\
Model architecture & ResNet-18 \\
Model variant & StoMPP-BNN \\
Optimizer & SGD (Nesterov momentum) \\
Learning rate & 0.1 \\
Momentum & 0.9 \\
Weight decay & 0 \\
LR schedule & Constant \\
Loss & Cross-entropy \\
Label smoothing & 0.0 \\
Masking scheme & Layerwise progressive masking \\
Mask schedule & Cubic ($p: 0 \rightarrow 1$) \\
Mask refresh rate & 100 \\
Activation function (unfrozen) & Clip \\
Frozen activation & Sign \\
Training epochs & 200 \\
Precision & FP32 \\
\bottomrule
\end{tabular}
\end{table}

\section{Training Curves of Variants}
\label{sec:appendix_b}

While we discuss the sawtooth-like training curves in Section~\ref{sec:training_dynamics} and Figure~\ref{fig:hp_sweep_and_curves} (b-d), we find that the hybrid architectures introduced in Section~\ref{sec:hybrids} and Figure~\ref{tab:compare_hybrids}, as well as BWNs, exhibit different training curves than STE or the standard layerwise StoMPP training algorithm.

\subsection{BWN Training Curves}

In the context of BWNs (Figure~\ref{fig:app_bwn_test}, Figure~\ref{fig:app_bwn_train}), we find that StoMPP shows a "swoop", initially increasing as it trains to the task, followed by a plateau or dip, and finally a region that accelerates tot the final accuracy again. We believe this may be because the network is harder to train as it is rapidly binarizing, and that performance and the representational capacity loss are "fighting" in the saddle. As the bulk of weights are trained, the network is mostly binary, and it can effectively learn the final portion of the network. This training curve behavior is of course dependent on the p value throughout training, and suggest further investigation into the nature of these curves.

\begin{figure}[H]
    \centering
    \includegraphics[width=0.9\linewidth]{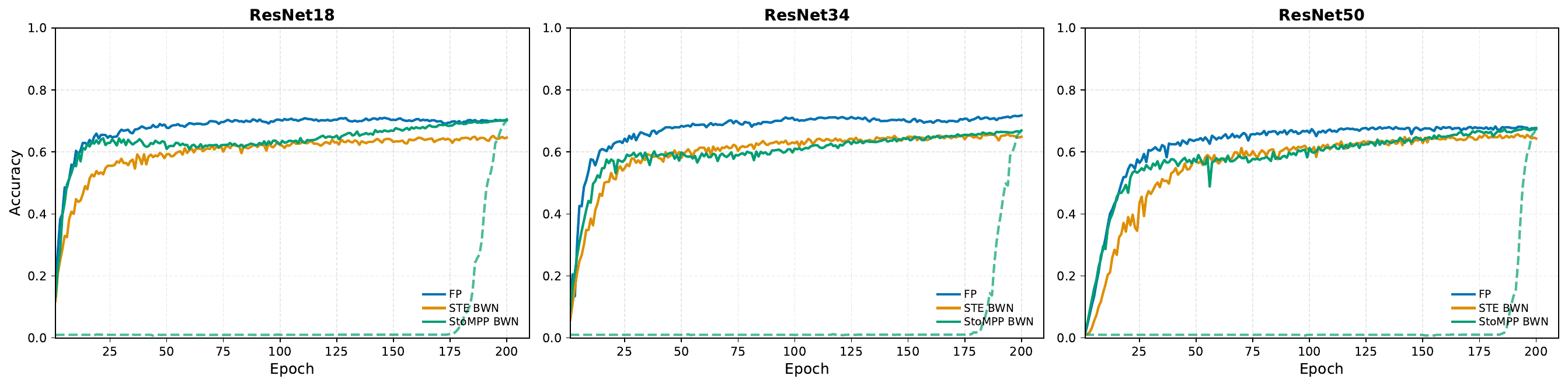}
    \caption{Testing Accuracy Curves for BWNs}
    \label{fig:app_bwn_test}
\end{figure}

\begin{figure}[H]
    \centering
    \includegraphics[width=0.9\linewidth]{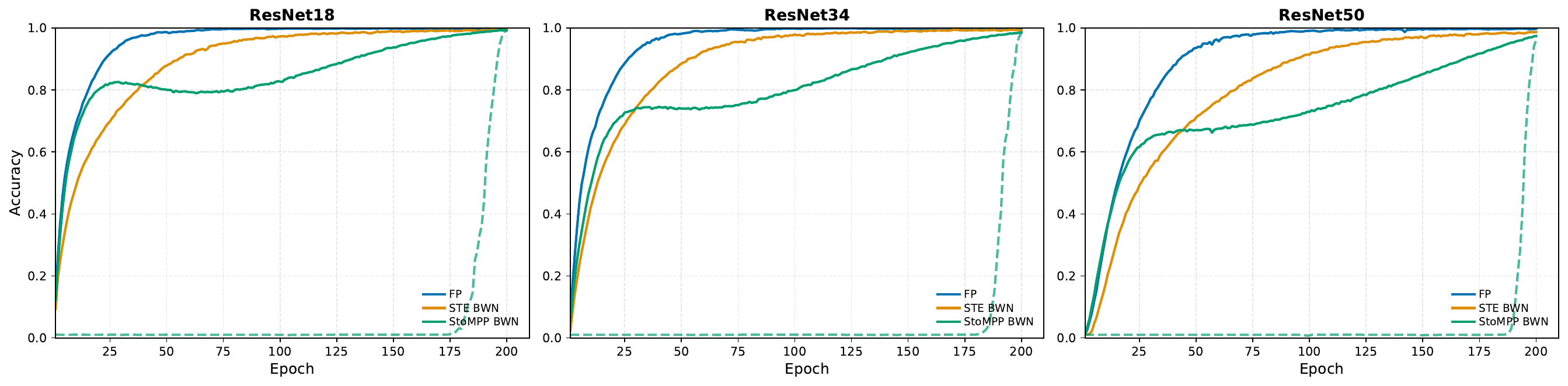}
    \caption{Training Accuracy Curves for BWNs}
    \label{fig:app_bwn_train}
\end{figure}

\subsection{StoMPP BNN Hybrids}

We observe that when Hybrid (StoMPP activations, STE weights) and Reverse Hybrid (StoMPP weights, STE activations) have significantly different training behavior beyond their final results. We find that while StoMPP has the strong sawtooth effect described as it learns each layer. In the context of a layerwise masking scheme, we find that reverse hybrid exhibits a sawtooth curve while hybrid does not. 

\begin{figure}[H]
    \centering
    \includegraphics[width=0.9\linewidth]{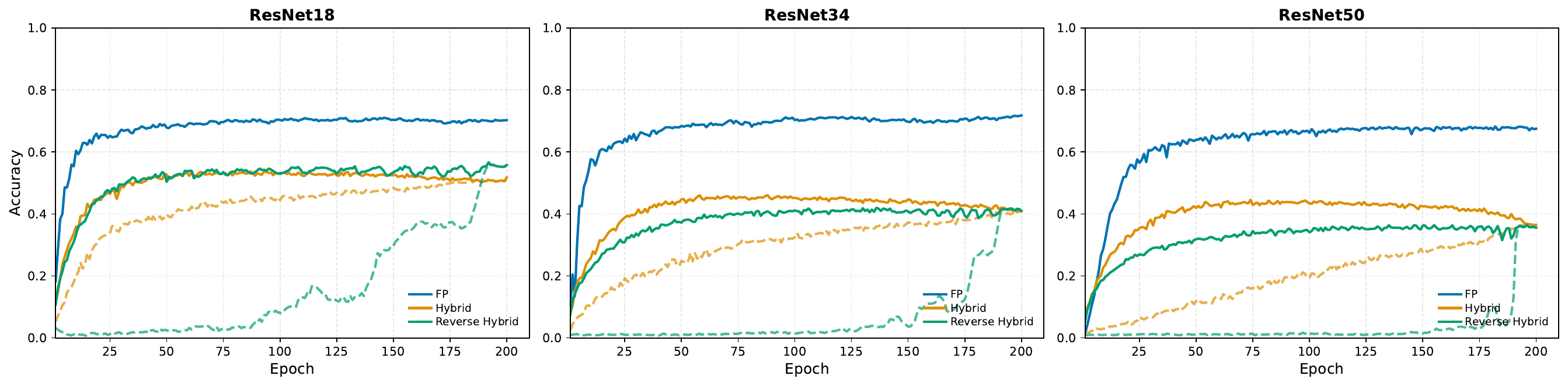}
    \caption{Testing Accuracy Curves StoMPP Hybrids}
    \label{fig:app_bwn_test}
\end{figure}

\begin{figure}[H]
    \centering
    \includegraphics[width=0.9\linewidth]{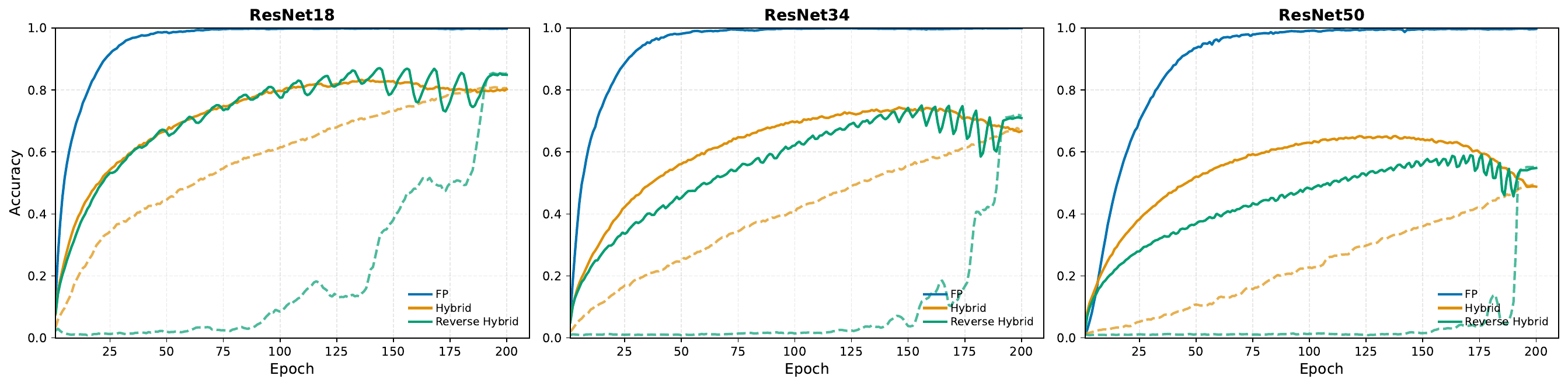}
    \caption{Training Accuracy Curves for StoMPP Hybrids}
    \label{fig:app_bwn_train}
\end{figure}

When applying the hybrid techniques using the global masking technique (which result in worse final performance, as shown in Section~\ref{sec:main_results}), we find that this sawtooth effect disappears. Based on the understanding that a global network mask has a plateau as $p$ rises, it makes sense that a layerwise mask applied this per layer. Despite the absence of a sawtooth effect for global masking, we still find that the convergence behavior of hybrid and reverse hybrid vary significantly. We find that reverse hybrid exhibits the same general contour as the BWN, with increasing accuracy as the freezing finalizes, while hybrid drops as the freezing finalizes. These effects are especially pronounced on the training accuracy since that is what the model is actually learning, although they are visible on the testing accuracy as well.

\begin{figure}[h!]
    \centering
    \includegraphics[width=0.9\linewidth]{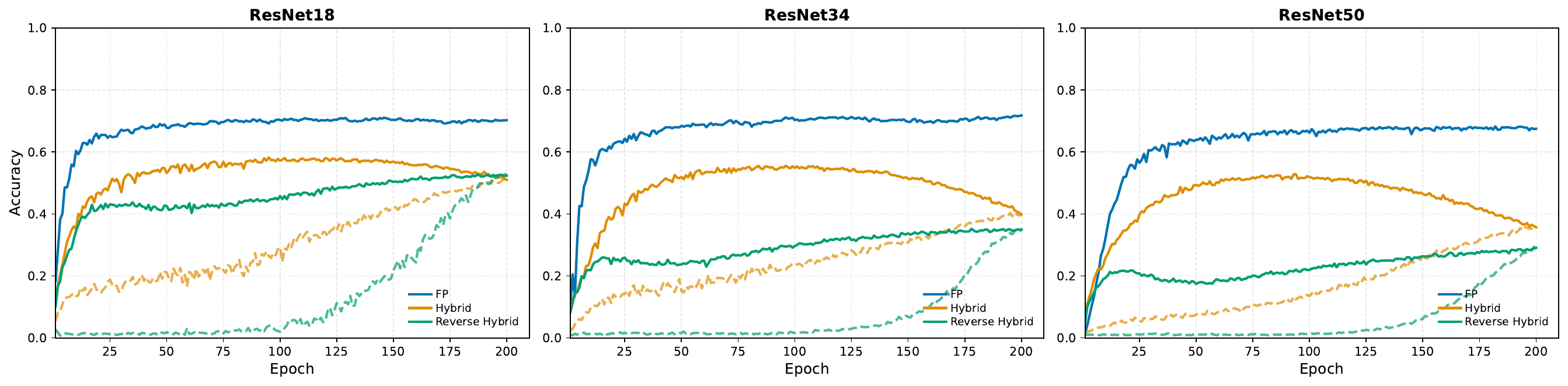}
    \caption{Testing Accuracy Curves for StoMPP Hybrids}
    \label{fig:placeholder}
\end{figure}

\begin{figure}[h!]
    \centering
    \includegraphics[width=0.9\linewidth]{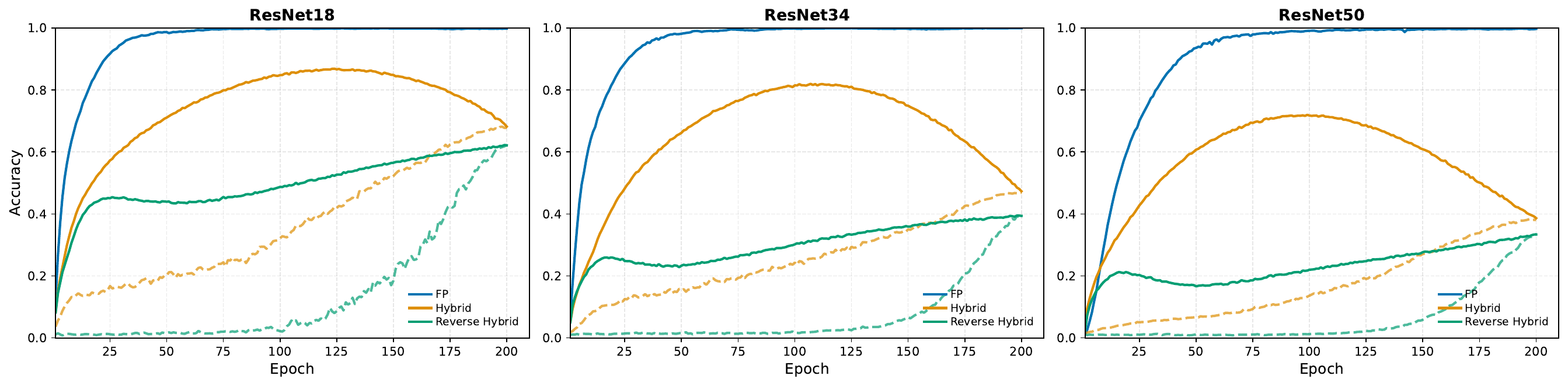}
    \caption{Training Accuracy Curves for StoMPP Hybrids}
    \label{fig:placeholder}
\end{figure}

\newpage
\subsection{Finetuning and BiReal-Net}

The training curves of BiReal-Net also show interesting behavior when combined with StoMPP. When applying layerwise masking in this context, we find that BiReal-Net significantly limits the sawtooth oscillations noted in Section~\ref{sec:training_dynamics}. We believe this is do to the additional residual connections added to the network, adding a signal as the activations and weights from the preceding layers are binarized. We also find that the quantized and unquantized version of stompp match much more closely throughout training, rather than just at the end, when training with BiReal net. The improved apparent stability of training StoMPP with BiReal network suggests possible architectural improvements may improve the ability of this technique. 

\begin{figure}[h!]
    \centering
    \includegraphics[width=0.7\linewidth]{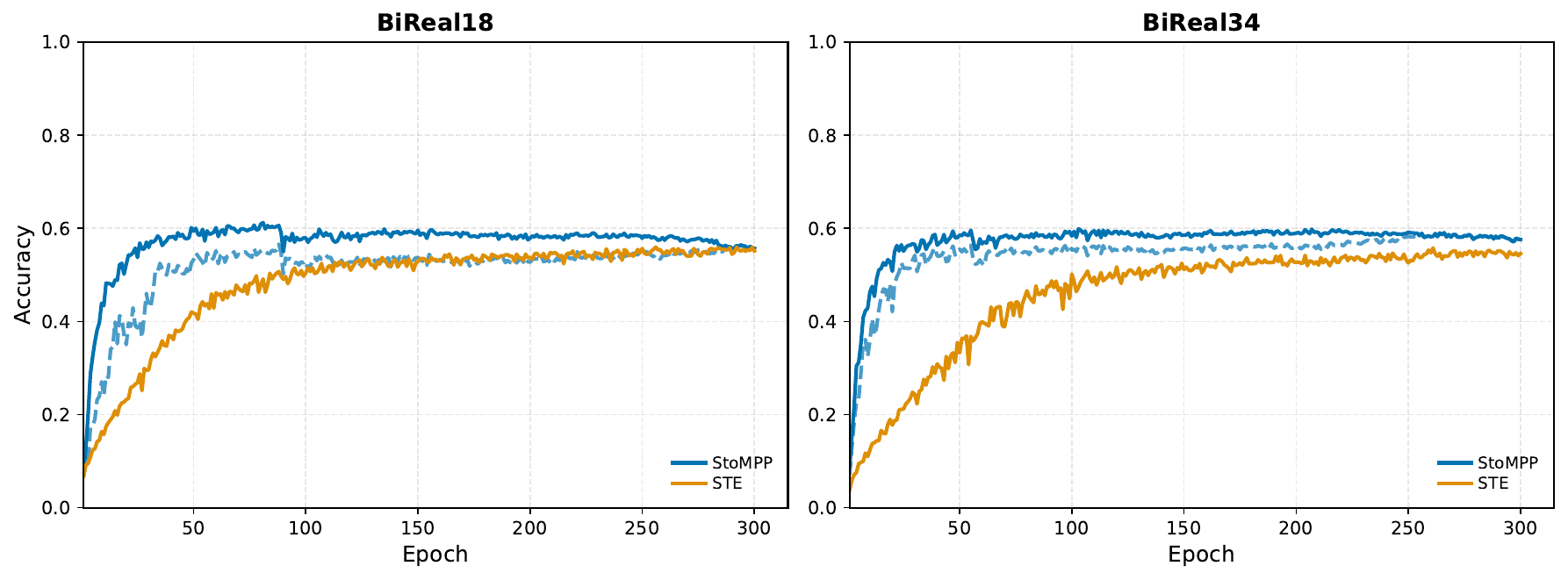}
    \caption{Testing Accuracy Curves for BiReal training from scratch}
    \label{fig:placeholder}
\end{figure}

\begin{figure}[h!]
    \centering
    \includegraphics[width=0.7\linewidth]{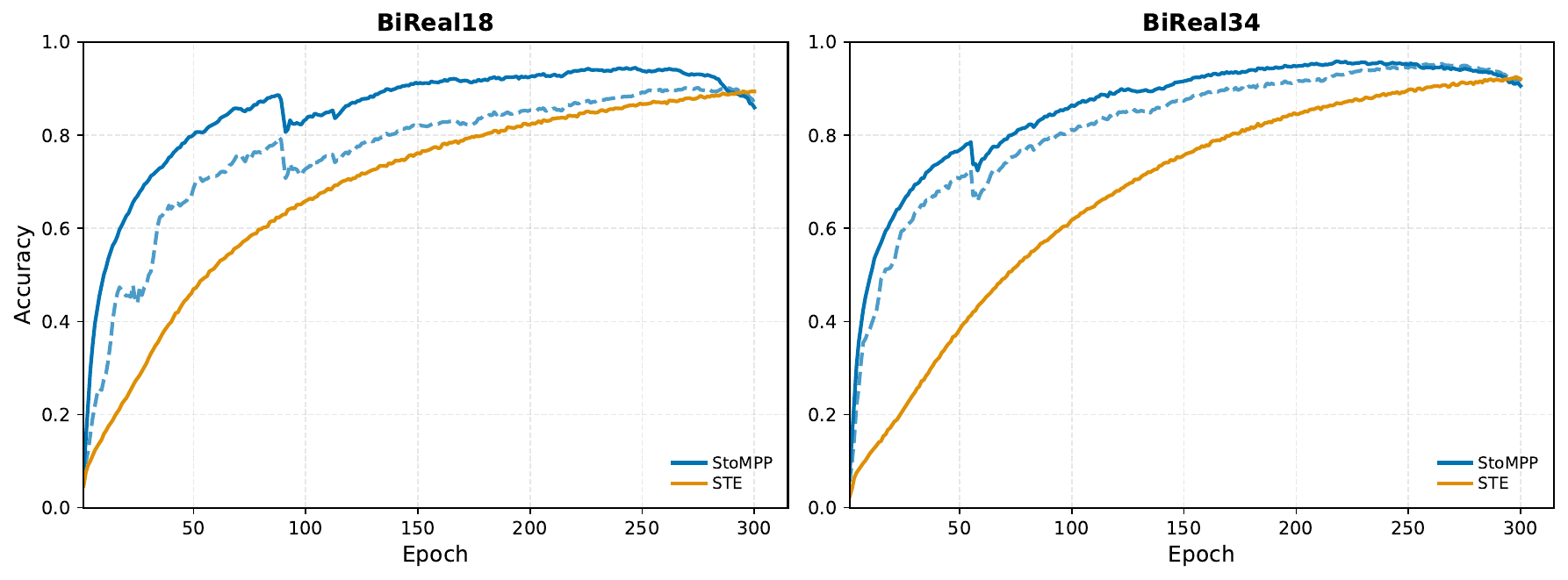}
    \caption{Training Accuracy Curves for BiReal training from scratch}   
    \label{fig:placeholder}
\end{figure}

We find also that training StoMPP/STE from a pretrained network, either in BiReal-Net or for typical ResNets, that both techniques are able to retain their performance. These networks maintain a generally stable 

\begin{figure}[h!]
    \centering
    \includegraphics[width=0.7\linewidth]{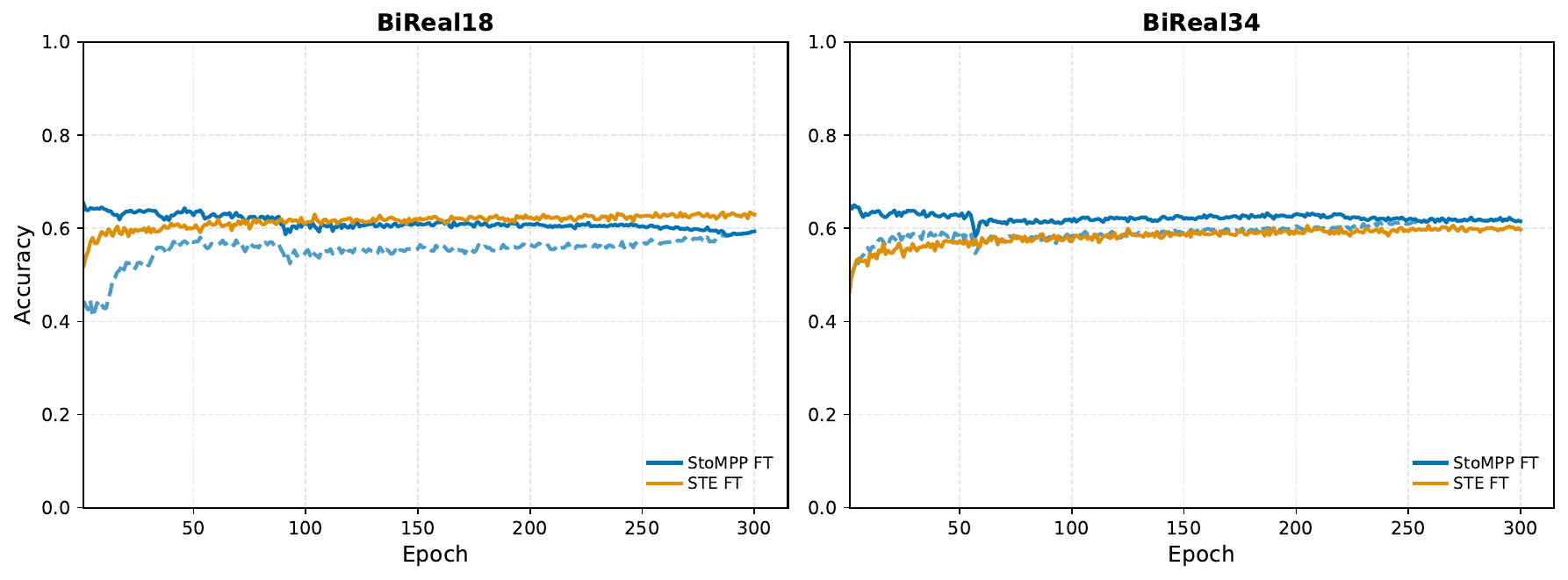}
    \caption{Testing Accuracy Curves for finetuned (FT) BiReal training}
    \label{fig:bireal_ft_test}
\end{figure}

\begin{figure}[h!]
    \centering
    \includegraphics[width=0.7\linewidth]{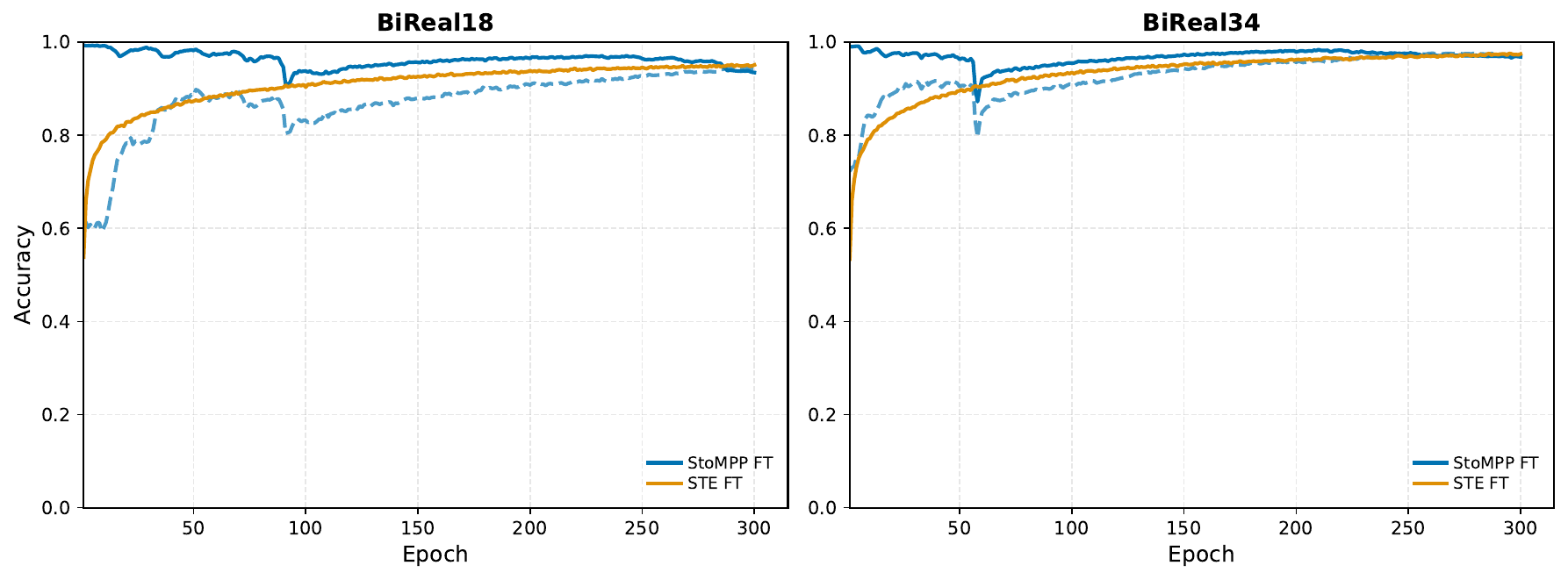}
    \caption{Training Accuracy Curves for finteunted (FT) BiReal training}   
    \label{fig:bireal_ft_train}
\end{figure}

\section{Hyperparameter Sweep of Forward Masking}
\label{sec:app_hp_forward_masking}

\begin{figure}[H]
    \centering
    \includegraphics[width=0.6\linewidth]{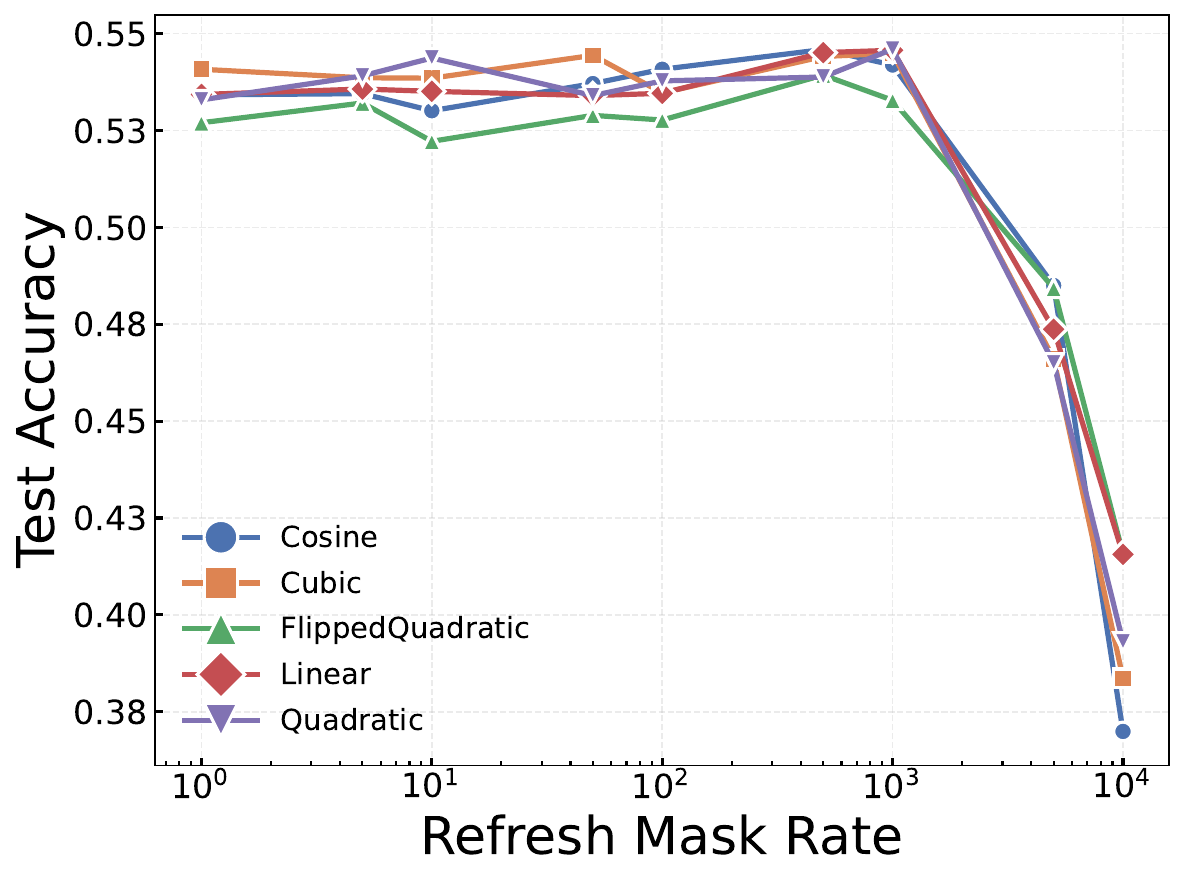}
    \caption{Hyperparameter Sweep of Masking Schemes under Layerwise Masking}
    \label{fig:app_hp_sweep_masking_schemes}
\end{figure}

We find that while the particular masking scheme is very important for the performance of a network when global masking is used, we find the structure of layerwise masking limits the effect of these hyperparameters (see Figure~\ref{fig:app_hp_sweep_masking_schemes}). Aside from applying layerwise masking, the apprach is identical to that of Figure~\ref{fig:hp_sweep_and_curves} (a), analyzing ResNet18 over 5 schedulers. See Section~\ref{sec:hp_ablations} for more details, and Appendix~\ref{sec:appendix_b} for exact training scheme specifics.

All schemes have similar performance, likely due to the positive layerwise effect limiting the interaction between the masking and gradient blockages. At high refresh rates, we find that performance still degrades in a similar manner to that of the hyperparameter search for global masking.

\section{Additional Related Work}
\label{sec:app_related_work}

We list additional binary neural network training methods and architectural variants not discussed in the main paper due to space constraints, acknowledging varied approaches and progress in extreme quantization.

\paragraph{Multiple Binary Bases and Approximation Methods.}
ABC-Net \cite{lin2017towards}, AdaBin \cite{tu2022adabinimprovingbinaryneural}, Projection CNN \cite{gu2018projectionconvolutionalneuralnetworks}.

\paragraph{Architectural Improvements.}
MeliusNet \cite{bethge2021meliusnet}, Group-Net \cite{zhuang2022structuredbinaryneuralnetworks}, BinaryDenseNet \cite{Bethge2019BinaryDenseNetDA}, Circulant Binary CNN \cite{liu2019circulantbinaryconvolutionalnetworks}.

\paragraph{Activation Function Design.}
ReActNet (PReLU-based) \cite{liu2020reactnet}, Unbalanced Activation \cite{kim2021improving}, Regularizing Activation Distributions \cite{ding2019regularizingactivationdistributiontraining}.

\paragraph{Optimizer and Training Strategies.}
AdamBNN \cite{liu2021adam}, Binary Optimizer (BOP) \cite{helwegen2019latentweightsexistrethinking}, SGDAT \cite{sgdat2023}.

\paragraph{Knowledge Distillation.}
Training with Knowledge Transfer \cite{leroux2020training}, Quantization-aware Knowledge Distillation (QKD) \cite{QKD_Kim_2019}, KDG-BNN \cite{gao2022memristive}.

\paragraph{Loss Functions and Regularization.}
Loss-Aware Binarization \cite{hou2017loss}, BinaryDuo \cite{kim2020binaryduoreducinggradientmismatch}, Defensive Quantization \cite{lin2019defensivequantizationefficiencymeets}.

% You can have as much text here as you want. The main body must be at most $8$ pages long.
% For the final version, one more page can be added.
% If you want, you can use an appendix like this one.  

% The $\mathtt{\backslash onecolumn}$ command above can be kept in place if you prefer a one-column appendix, or can be removed if you prefer a two-column appendix.  Apart from this possible change, the style (font size, spacing, margins, page numbering, etc.) should be kept the same as the main body.
%%%%%%%%%%%%%%%%%%%%%%%%%%%%%%%%%%%%%%%%%%%%%%%%%%%%%%%%%%%%%%%%%%%%%%%%%%%%%%%
%%%%%%%%%%%%%%%%%%%%%%%%%%%%%%%%%%%%%%%%%%%%%%%%%%%%%%%%%%%%%%%%%%%%%%%%%%%%%%%

\end{document}